\documentclass{article}
\usepackage{iclr,times}
\iclrfinalcopy
\iclrfinalcopy
\usepackage[breakable, skins, xparse]{tcolorbox}
\usepackage{microtype}
\usepackage{graphicx}
\usepackage{subfigure}
\usepackage{booktabs} % for professional tables
\usepackage{enumitem}
\usepackage{hyperref}

\usepackage{amsmath}
\usepackage{amssymb}
\usepackage{mathtools}
\usepackage{amsthm}

\usepackage[capitalize,noabbrev]{cleveref}

\theoremstyle{plain}

\theoremstyle{definition}

\theoremstyle{remark}

\usepackage[textsize=tiny]{todonotes}
\usepackage{xcolor}
\usepackage{url}
\usepackage{arydshln}
\usepackage{soul}

\usepackage[utf8]{inputenc} % allow utf-8 input
\usepackage[T1]{fontenc}    % use 8-bit T1 fonts
\usepackage{booktabs}       % professional-quality tables
\usepackage{amsfonts}       % blackboard math symbols
\usepackage{nicefrac}       % compact symbols for 1/2, etc.
\usepackage{microtype}      % microtypography
\usepackage{xcolor}         % colors
\usepackage{microtype}
\usepackage{graphicx}
\usepackage{tabularx}
\usepackage{subfigure}
\usepackage{booktabs}
\usepackage{hyperref}
\usepackage{color}
\usepackage{xcolor}
\usepackage{amsmath}
\usepackage{amssymb}
\usepackage{mathtools}
\usepackage{amsthm}
\usepackage[capitalize,noabbrev]{cleveref}
\usepackage{algorithm}
\usepackage{algorithmic}
\usepackage[textsize=tiny]{todonotes}
\usepackage{multirow}
\usepackage{float}
\usepackage{stfloats}
\usepackage{times}
\usepackage{epsfig}

\usepackage{wrapfig}
\usepackage{enumitem}

\usepackage{subfiles}

\usepackage{color}
\usepackage{colortbl}
\setlist[itemize]{leftmargin=2em}

\usepackage{wrapfig}
\usepackage{makecell}   % for \makecell (line breaks in header)
\usepackage{subfigure} 
\usepackage{amssymb}
\usepackage{amsbsy}
\usepackage{wrapfig}
\usepackage{bbding}
\usepackage{color}
\usepackage{fancyhdr}

\renewcommand{\algorithmicrequire}{\textbf{Input:}}
\renewcommand{\algorithmicensure}{\textbf{Output:}}
\newcommand{\model}{DeepInnovator}

\newcommand{\qwenplus}{Qwen-plus}
\newcommand{\qwenit}{Qwen-14B-Instruct}

\usepackage{wrapfig}

\title{\model: Triggering the Innovative Capabilities of LLMs }

\author{ Tianyu Fan$^{\clubsuit}$, Fengji Zhang$^{\spadesuit}$, Yuxiang Zheng$^{\diamondsuit}$, {Bei Chen}$^{\heartsuit}$, Xinyao Niu$^{\heartsuit}$,\\ ~\textbf{Chengen Huang}$^{\heartsuit}$, \textbf{Junyang Lin}$^{\heartsuit}$, \textbf{Chao Huang}$^{\clubsuit}$\\
$^\clubsuit$The University of Hong Kong, $^\spadesuit$City University of Hong Kong\\
$^\diamondsuit$Shanghai Jiao Tong University, $^\heartsuit$Alibaba Group \\
}

\begin{document}
\maketitle

\begin{abstract}
    The application of Large Language Models (LLMs) in accelerating scientific discovery has garnered increasing attention, with a key focus on constructing research agents endowed with innovative capability, i.e., the ability to autonomously generate novel and significant research ideas. Existing approaches predominantly rely on sophisticated prompt engineering and lack a systematic training paradigm. To address this, we propose \model, a training framework designed to trigger the innovative capability of LLMs. 
    Our approach comprises two core components.  
    (1) \emph{``Standing on the shoulders of giants''}. We construct an automated data extraction pipeline to extract and organize structured research knowledge from a vast corpus of unlabeled scientific literature.  
    (2) \emph{``Conjectures and refutations''}. We introduce a ``Next Idea Prediction'' training paradigm, which models the generation of research ideas as an iterative process of continuously predicting, evaluating, and refining plausible and novel next idea.
Both automatic and expert evaluations demonstrate that our \textbf{\model-14B} significantly outperforms untrained baselines, achieving win rates of 80.53\%–93.81\%, and attains performance comparable to that of current leading LLMs. This work provides a scalable training pathway toward building research agents with genuine, originative innovative capability, and will open-source the dataset to foster community advancement.
Source code and data are available at:
\textcolor{blue}{\url{https://github.com/HKUDS/DeepInnovator}}.
\end{abstract}

\newcommand{\sw}{preceding context}
\newcommand{\xw}{following context}

\section{Introduction}
\label{sec:intro}

Scientific discovery is fundamentally a process of building upon prior milestones. As Isaac Newton famously articulated, \emph{``If I have seen further, it is by standing on the shoulders of giants''} \citep{newtoncorrespondence}. This insight suggests that the birth of a new idea is not an isolated event but a logical progression rooted in an existing body of work. To act as an effective innovator, one must systematically analyze how prior studies relate to each other by identifying shared objectives, conflicting results, and remaining gaps. This synthesis is essential for moving beyond incremental changes toward meaningful discovery.

However, a nascent idea is merely the beginning. The growth of scientific knowledge proceeds through \emph{``conjectures and refutations''} \citep{popper2014conjectures}. Innovation is not a one-off event but a process of iterative refinement. An innovator must not only synthesize existing literature but also possess the capability to critique its own proposals and leverage feedback to transform a rough concept into a crystallized research direction.

Large Language Models (LLMs) have recently demonstrated significant potential in the scientific domain. Existing agents aid in research tasks such as literature surveys~\citep{xu2025comprehensive}, manuscript writing~\citep{tang2025ai}, or code generation~\citep{weng2025deepscientist,novikov2025alphaevolve}. However, a more fundamental question lies in the conceptualization phase: \emph{Can LLMs autonomously generate novel and significant research ideas?}~\citep{si2024can}. 
% To systematically explore this capability, we investigate three underlying problems: (1) how to effectively leverage vast corpora of unannotated scientific papers to structure existing knowledge; (2) how to design a generation pipeline that fosters the iterative refinement of high-quality ideas; and (3) how to reliably assess the novelty and feasibility of the generated ideas.
To systematically explore this innovative capability, we address two fundamental problems in training research agents: (1) how to effectively leverage vast corpora of unannotated scientific papers to structure existing knowledge; and (2) how to design a training pipeline that stimulates the research agent’s capability for innovation, enabling it to autonomously generate and refine research ideas.

In this paper, we explore these problems through a clearly defined task: given a set of existing studies, determine whether LLMs can produce a logically coherent and directionally sound research idea. We propose \model, an agent training framework designed to mimic the dual processes of idea generation and refinement. The framework consists of two main components. The first is \textbf{Automated Data Extraction and Synthesis}. This module extracts key insights and inter-paper relationships from a vast corpus of unannotated scientific papers to construct a comprehensive research context. This process effectively allows the model to \emph{``stand on the shoulders of giants''} by grounding generation in a structured understanding of prior work. The second component is a \textbf{``Next Idea Prediction'' } training task. This approach requires the research agent to iteratively generate an improved idea based on the previous version and a generated comment, thereby eliciting the agent’s innovative capability. We propose \model-14B, a research agent built upon \qwenit, via Reinforcement Learning (RL), utilizing process rewards to assess incremental improvements while employing a decomposition mechanism to prevent reward hacking. This iterative optimization directly embodies the cycle of \emph{``conjectures and refutations''}, enabling the agent to self-correct and refine its research idea.

To evaluate our approach, we conducted both automatic comparisons and expert evaluation. Our \model-14B~surpasses an 80$\%$ win rate against baselines across four distinct dimensions. Furthermore, despite being trained specifically on papers from mathematics, finance, statistics, and computer science, \model~demonstrates strong generalization capabilities. It generates research ideas in out-of-distribution fields, such as law and biotechnology, that significantly outperform those produced by the untrained base model and occasionally surpass GPT-4o. These results indicate that our carefully designed training methodology successfully enhances the innovative capability of research agents. To foster further research in scientific AI, we publicly release our code and data to the community.
\section{Related Works}
\textbf{Agents for Research.}
The application of LLMs in scientific research has evolved from passive literature synthesis toward active ideation, yet a critical capability gap persists. Current systems primarily operate as \textit{DeepResearch} assistants that orchestrate multi-step workflows to exhaustively retrieve and compress existing knowledge~\citep{xu2025comprehensive, openaidr, zheng2025deepresearcher, zhang2025deep}. While effective at reducing cognitive load in literature review, these approaches fundamentally function as knowledge compressors that summarize what has already been done~\citep{fan2025understanding, jin2025search}. Efforts in automated scientific discovery fall into two categories: domain-specific solvers like AlphaFold~\citep{jumper2021highly} and GNoME~\citep{merchant2023scaling}, which optimize algorithms toward predefined objectives within their specialized domains;
and end-to-end research agents such as AI Scientist~\citep{yamada2025ai} and AI-Researcher~\citep{tang2025ai} that automate workflows from ideation to manuscript writing but often prioritize engineering execution over ideation quality, relying on code-based validation that limits applicability to non-computational domains. 
They all represent implementations of a good research idea. Here, we aim to discuss a more general research agent designed for the generation of high-quality research ideas.
% Crucially, both paradigms remain anchored in \textit{information synthesis} rather than genuine \textit{scientific ideation}—the ability to formulate novel, logically grounded hypotheses beyond existing literature.

\textbf{RL in Open-ended Domains.}
Reinforcement learning offers a pathway toward ideation but faces fundamental challenges in open-ended scientific domains. While RL has succeeded in deterministic settings with objective rewards (e.g., mathematics~\citep{shao2024deepseekmath, deepseekr1} and code generation~\citep{fan2025posterior}), scientific innovation lacks oracle verifiers, necessitating subjective evaluation via LLM-as-a-Judge~\citep{lee2024rlaif, maeureka} or rubric-based frameworks~\citep{gunjal2025rubrics, viswanathan2025checklists}. These approaches, however, are vulnerable to reward hacking where models exploit surface-level judge preferences~\citep{huang2025reinforcement, shao2025dr, sharma2024critical}. 
We design a method that decouples reward and comment, separating outcome scoring and improvement suggestions to suppress reward hacking and encourage the generation of intrinsically valuable ideas.
% \model~bridges these gaps through two innovations: (1) a \textit{Next Idea Prediction} paradigm that trains agents to forecast the ``research delta''—the logical leap from current knowledge to future breakthroughs—shifting focus from summarizing the past to predicting novel hypotheses; and (2) a \textit{process-oriented delta reward} with decoupled evaluation and feedback, separating outcome scoring from improvement instructions to discourage reward hacking and encourage intrinsically valuable ideation. By learning patterns of scientific evolution without dependence on code execution sandboxes, \model~generalizes across disciplines while prioritizing ideation quality over workflow automation.
\section{Problem Formulation for Innovation task}

Many different ``innovation tasks'' have been undertaken by the community in the domain of building research agent.
In existing work, different systems embody this concept in varying ways: AI-Researcher~\citep{tang2025ai} and AI-Scientist~\citep{yamada2025ai} define innovation tasks as having AI research assistants write research papers, using paper quality as a measure of innovative capability; whereas DeepScientist~\citep{weng2025deepscientist} and AlphaEvolve~\citep{novikov2025alphaevolve} assess innovation by having AI assistants write code that improves algorithmic performance in specific domains. Despite these differing forms, such tasks fundamentally involve the realization and development of a research idea. 
Here, we aim to explore a more general and fundamental form of innovation task: \textit{how can foundation models be trained to autonomously generate high-quality research ideas?}

We formulate the innovation task as a sequential idea generation and refinement process: a base LLM acts as a research agent that, given contextual information $C$ from a set of reference papers, autoregressively generates an initial research idea $y^{(0)}$ and iteratively refines it into increasingly sophisticated formulations. The training objective is to predict the ``next improved idea'' $y^{(i+1)}$ at each step $i$.

Formally, let \( \mathcal{R} = \{r_1, r_2, \dots, r_N\} \) denote a collection of reference papers. Their content is encoded into an initial context \( C = \text{Encode}(\mathcal{R}) \), which serves as the starting state for the generation process. The agent’s policy \( \pi_\theta \)  operates autoregressively over a sequence of tokens, producing a trajectory of progressively refined ideas.

We define the action space to consist of a single generative action <idea></idea>, which generates a segment of idea text. The full generation is partitioned into a sequence of idea snapshots \( y^{(0)}, y^{(1)}, \dots, y^{(K)} \), where \( y^{(0)} \) is the initial, coarse-grained research idea, each subsequent \( y^{(k)} \) (for \( k \geq 1 \)) represents a semantically complete and improved version of the idea, obtained by continuing token generation from the previous state.
The final output \( y^{(K)} \) is treated as the model’s proposed innovative research idea, where \( K \) denotes the number of iterations. Critically, this idea is not a direct paraphrase or extraction from \( \mathcal{R} \), but rather emerges through an internal iterative refinement loop, where each new idea is conditioned on both the reference context and the model’s own prior outputs, effectively simulating self-critique and creative evolution.

 \begin{figure*}[t]
    \begin{center}
    % \vspace{-2.5em}
    \includegraphics[width=0.8\textwidth]{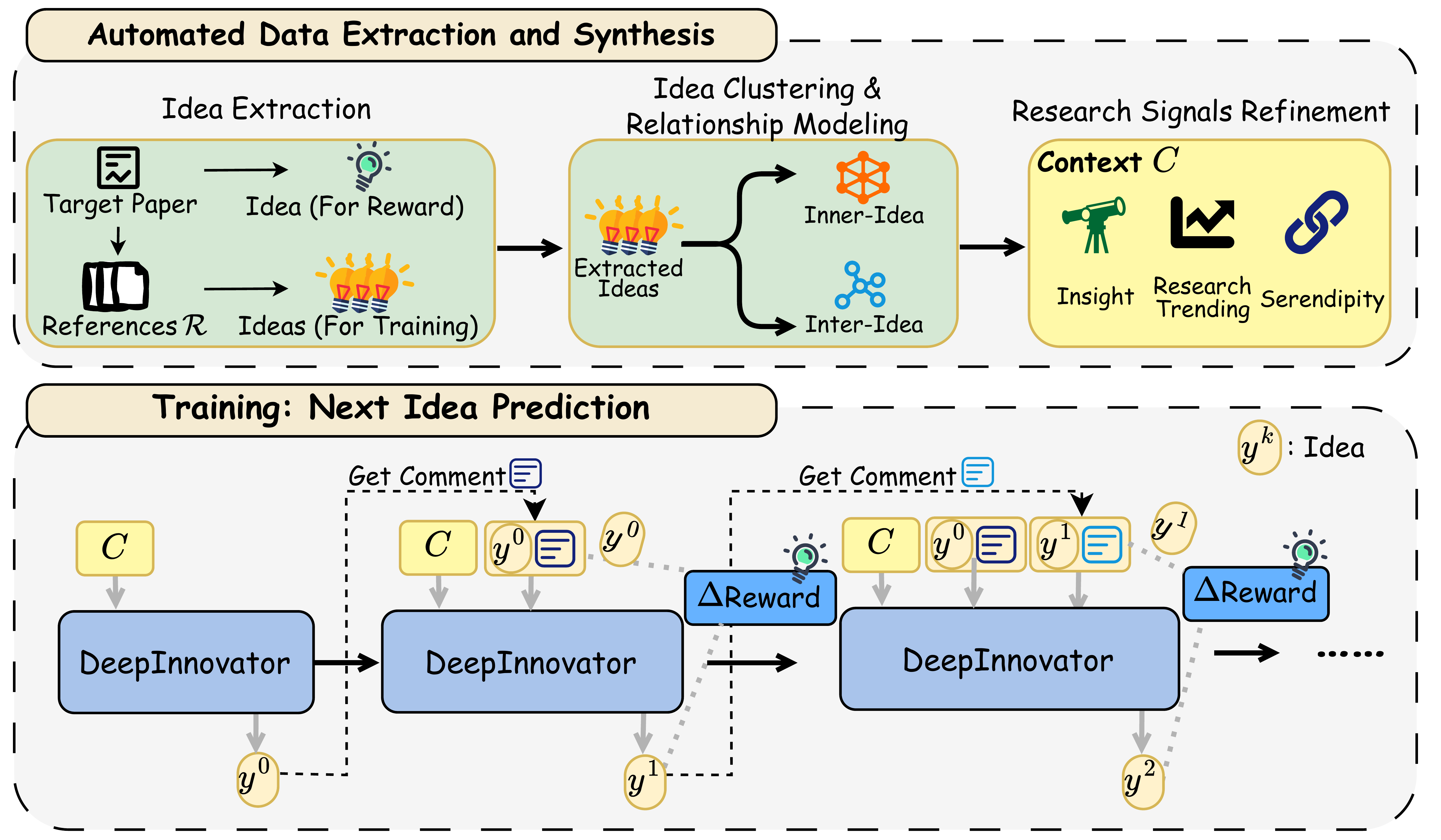}
    \end{center}
    \vspace{-1em}
    \caption{The \model\ Framework. Top: We construct training data from arXiv through a carefully designed automated data extraction and synthesis pipeline (Sec.~\ref{sec:data}).  
Bottom: We perform RL training via a meticulously designed next-idea prediction task (\ref{sec:train_obj}), coupled with a decoupled reward and critique mechanism (Sec.~\ref{sec:reward}).
    }
    % \vspace{-1em}
    \label{Fig:model}
    % \vspace{-0.5em}
\end{figure*}

\section{Training Methodology of \model}
\label{sec:solution}

A fundamental challenge in constructing research agents capable of supporting scientific discovery lies in how to transform the innovation potential embedded within LLMs, which is derived from human knowledge corpora, into explicit behaviors that are guidable, evaluable, and continuously optimizable.
This challenge gives rise to three interrelated training challenges:

 \textbf{Challenge 1: Lack of Training Data.} Unlike tasks in domains with well-defined correct answers, such as code generation, search, or mathematical reasoning, innovation tasks lack well-defined training data, readily available training objectives, and clear correctness criteria.

\textbf{Challenge 2: Inapplicability of Predefined Objectives in Research Tasks.}  
Previous approaches to designing research agents~\citep{tang2025ai,weng2025deepscientist,novikov2025alphaevolve} typically rely on predefined objectives or hard-coded validation logic. However, formulating a well-defined research task presupposes the existence of a sound research idea. When the research idea remains ambiguous and the validation pathway is not yet established, such methods are often ineffective.

 \textbf{Challenge 3: Reward Hacking in Open-Ended Evaluation.}
 In such unbounded environments, relying solely on LLM-as-a-Judge to provide rewards is highly prone to reward hacking~\citep{Gao2023ScalingLF, Zhao2025OneTT,Hu2025BiasFT}, 
 % particularly incentivizing sycophantic responses that cater to the judger’s presumed preferences rather than producing objective insights~\citep{Anthropic2025NaturalEM}.
 particularly incentivizing sycophantic responses that cater to the judge’s presumed preferences rather than delivering insightful analysis of substantive value~\citep{Anthropic2025NaturalEM}.

To address these challenges, we propose the following solutions:  
\textbf{(1)} We meticulously design a method for extracting training data from a large corpus of unannotated scientific papers (Sec.~\ref{sec:data});  
\textbf{(2)} We formulate the Next Idea Prediction task to train \model (Sec.~\ref{sec:train_obj});  
\textbf{(3)} We introduce a decoupled architecture that separates rewards from comments to prevent reward hacking (Sec.~\ref{sec:reward}).

\subsection{Automated Data Extraction and Synthesis}\label{sec:data}

To bolster the innovative capabilities of agents in research tasks, we design an automated data synthesis pipeline to construct the ``Next Idea Prediction'' training task from arXiv. 
As shown in Fig.~\ref{Fig:model}, the objective of this task is as follows: given all references $\mathcal{R}$ of a target paper, the research agent must generate and iteratively refine the next research idea $y$ based on its own knowledge and the context $C$ composed of information from these reference papers. This setup simulates the cognitive process of performing knowledge integration and leapfrogging innovation in real-world scientific research.
Specifically, for each target paper, we first automatically collect its references $\mathcal{R}$ and treat these references as the prior knowledge context. However, directly using the original reference paper texts presents two major challenges: (1) papers are often lengthy, and combining multiple references easily exceeds the agents' context window limits; and (2) the original texts contain substantial redundant details, making it difficult to highlight the relationships among key research insights. 

To address this, we introduce a hierarchical abstraction pipeline that transforms the original citations into compact, structured representations of research ideas.
Our pipeline comprises the following three key steps:

% \begin{enumerate}
    \textbf{Idea Extraction}:  
    % Using carefully designed prompts, we extract the core research ideas from each cited paper and formalize them into structured statements. This step significantly compresses information density while preserving semantic completeness.
Using carefully designed prompts, we extract the research idea from the target paper as the real idea used by the reward model. Additionally, we extract the core research ideas from each cited paper of the target paper and formalize them into structured statements. This step significantly compresses information density while preserving semantic completeness.

    \textbf{Idea Clustering and Relationship Modeling}:  
    We perform semantic clustering on all extracted ideas and identify two types of critical relationships:
    \begin{itemize}[leftmargin=*]
        \item \textit{Inner-idea relations}: Evolutionary, variant, or refinement pathways among ideas within the same cluster;
        \item \textit{Inter-idea relations}: Cross-cluster interactions such as integration, convergence, or conflict between ideas from different clusters.
    \end{itemize}
    This structured organization not only reveals the knowledge topology of prior work but also provides a logical scaffold for subsequent insight generation.

    \textbf{Higher-Order Research Signal Refinement}:  
    Building upon the idea relationships, we further distill three categories of higher-order research signals:
    \begin{itemize}[leftmargin=*]
        \item \textit{Insight}: Non-obvious patterns or contradictions inferred from multiple citation ideas (e.g., ``existing methods neglect temporal consistency in cross-modal alignment'');
        \item \textit{Research Trending}: Emerging or declining research directions identified through progressive relationships among ideas;
        \item \textit{Serendipity}: Latent connections between seemingly unrelated domains (e.g., ``transferring exploration mechanisms from reinforcement learning to neural architecture search'').
    \end{itemize}
% \end{enumerate}
We explicitly model these three types of signals not only to circumvent context-length limitations but, more importantly, because they correspond to three fundamental cognitive mechanisms in human scientific reasoning:  
\textit{inductive inference} (Insight)~\citep{newell2001theory,harvey2014creative},  
\textit{prospective judgment} (Research Trending)~\citep{chen2006citespace,min2021identifying}, and  
\textit{cross-domain association} (Serendipity)~\citep{kennedy2022serendipity,kang2022augmenting,gentner2011analogy}.  
These patterns are recognized as core drivers of breakthrough innovation.
By formalizing these cognitive primitives into computable intermediate representations, we provide the research agent with a scaffold that emulates human-like scientific thinking—thereby allowing it to generate innovative and forward-looking predictions rather than simply reproducing existing content.

% \subsection{Training Objective of \model}

\subsection{Reward Signal Design}
\label{sec:train_obj}

Inspired by recent work on contrastive delta rewards in foundation LLM training~\citep{seed2025seed1} and Long-Horizon Agents~\citep{wang2025harnessing}, we design a \textbf{process-oriented delta reward} mechanism in the scientific domain to enhance the model's innovative capabilities in environments lacking ground-truth answers and predefined training objectives.

This mechanism focuses not only on whether the final generated research idea approximates the actual research idea, but more critically on quantifying the magnitude of improvement introduced at each refinement step.
In fact, scientific breakthroughs rarely occur instantaneously; they typically emerge through iterative cycles of trial and error, reflection, and recombination. The core idea is that rewards should reflect the cognitive progress demonstrated by the agent throughout the iterative process. Through this process-oriented reinforcement signal, the agent is encouraged to engage in exploratory reasoning.

Specifically, we optimize the research agent's policy \(\pi_\theta\) within the Group Relative Policy Optimization (GRPO)~\citep{shao2024deepseekmath} to enable it to generate multi-turn research ideas that exhibit sustained cognitive progress, conditioned on a given citation context \(C\). For each context \(C\), we sample \(G\) response groups, with each group containing a complete trajectory \(o_i = (y_i^{(0)}, y_i^{(1)}, \dots, y_i^{(K)})\). The scalar reward for this trajectory is obtained by accumulating stepwise improvement signals provided by an external reward LLM \texttt{Reward}. 

During training, at step \( k \), the \texttt{Reward} module simultaneously observes the currently generated idea \( y^k \), the previously generated idea \( y^{k-1} \), and the real idea, and assigns a score to each of the two generated ideas; the difference between these scores serves as the reward for the current step.
The final reward $R(o_i)$ can be formulated as:
\begin{equation}
R(o_i) = \sum_{k=1}^{K} \texttt{Reward}\big(y_i^{(k-1)}, y_i^{(k)}; q\big),
\end{equation}

Based on this, we subsequently compute the normalized intra-group advantage estimate \(\hat{A}_i\). The final GRPO loss function is:
\vspace{-1em}
% \begin{equation}
% \mathcal{L}_{\text{GRPO}}(\theta) = -\mathbb{E}_{q \sim P(Q)} \Bigg[ \frac{1}{G} \sum_{i=1}^G \frac{1}{|o_i|} \sum_{t=1}^{|o_i|} \min\left( 
% r_{i}(\theta) \hat{A}_i, \ 
% \text{clip}\left(r_{i}(\theta), 1-\varepsilon, 1+\varepsilon\right) \hat{A}_i 
% \right) + \beta \, \text{KL}\Big(\pi_\theta \big\| \pi_{\text{ref}}\Big) \Bigg].
% \end{equation}
\begin{multline}
\mathcal{L}_{\text{GRPO}}(\theta) = -\mathbb{E}_{q \sim P(Q)} \Bigg[ \frac{1}{G} \sum_{i=1}^G \frac{1}{|o_i|} \sum_{t=1}^{|o_i|} 
\min\left( 
r_{i,t}(\theta) \hat{A}_i, \ 
\text{clip}\left(r_{i,t}(\theta), 1-\varepsilon, 1+\varepsilon\right) \hat{A}_i 
\right) \\
+ \beta \, \text{KL}\Big(\pi_\theta \big\| \pi_{\text{ref}}\Big) \Bigg].
\end{multline}

where \(|o_i|\) denotes the trajectory length, the advantage estimate \(\hat{A}_i\) is derived from the cumulative improvement score \(R(o_i)\), \( r_{i,t}(\theta) = \dfrac{\pi_\theta(a_{i,t} \mid s_{i,t})}{\pi_{\text{old}}(a_{i,t} \mid s_{i,t})} \) is the probability ratio of the new and old policies selecting action \( a_{i,t} \) under state \( s_{i,t} \), used to measure the magnitude of the policy update, and \(\text{clip}(\cdot)\) constrains the policy update step size to prevent excessive deviation from the old policy.

By aligning the reward design with next-idea prediction through a progressive, process-oriented formulation, our approach avoids reliance on a single ``ground-truth'' (which is quite difficult to get in the innovation task) and instead emphasizes the evolution of the cognitive process to unlocking the innovative potential of LLMs.

\subsection{Decomposing Reward and Comment in Open-ended Tasks}\label{sec:reward}

In RL frameworks that rely solely on reward models, agents may hack the reward by generating text that superficially exhibits high-scoring characteristics but actually deviates from human intent. 
To address this, we introduce text-guided improvement directions (comment) that explicitly identify, at the semantic level, the issues with the current idea and specify the desired correction, thereby helping the \model~avoid reward hacking. We design a decoupled architecture that separates reward and comment, explicitly partitioning the evaluation process into two independent components: the research agent receives improvement suggestions from the comment model \texttt{Comment}, but whether an improvement is deemed effective is determined solely by the reward model.

Specifically, during training, for each draft idea $y^{(k)}$ generated by the agent, we invoke a comment LLM $\texttt{Comment}$ to analyze its performance within the context of relevant literature $C$, and output structured suggestions—for example: ``fails to address temporal consistency'' or ``attempts to solve a problem already addressed in existing literature''. The agent then uses this feedback from $\texttt{Comment} $ to produce an improved version $y^{(k+1)}$, thereby simulating an authentic peer-review iteration process.

By decoupling process guidance from outcome evaluation, our design enforces a strict separation of responsibilities: the comment model ${\texttt{Comment} }$ instructs how to improve while the reward model $\texttt{Reward} $ solely judges whether it has genuinely improved. This effectively blocks common reward hacking pathways. 
% The trained research agent cannot please the reward model by echoing the phrasing of comments, inserting keywords that trigger high scores, or padding explanations with verbose but vacuous content, because the comment model ${\texttt{Comment} }$, which provides feedback to the agent—is entirely excluded from the reward computation. 
Since the reward signal is entirely independent of the comment model ${\texttt{Comment}}$, the trained research agent cannot please the reward mechanism by mimicking the phrasing of comments, inserting high-scoring keywords, or padding responses with verbose yet vacuous content, as the comment model ${\texttt{Comment}}$, which provides feedback to the agent, is completely excluded from the reward computation.
Consequently, the agent is compelled to focus on substantively improving the quality of the idea itself, rather than fabricating an illusion of progress.

% \vspace{-0.5em}
\section{Experiments}

\subsection{Experimental Settings}
\label{exp_setting}
\textbf{Datasets.}
Using the data collection process described in Sec.~\ref{sec:data}, we constructed the training dataset for \model. It comprises target papers extracted from four domains on arXiv, along with all their references and the original ideas of the target papers. We only collected papers published after March 2025 to mitigate data leakage issues. 
We curated a training set containing 1,012 research ideas and a validation set containing 113 research objectives, covering Computer Science, Mathematics, Finance, and Statistics. 

% We have released this dataset and provide examples from all domains in Appendix.

% \textbf{Implementation Details.}
% \begin{itemize}[leftmargin=*]
     \textbf{Model.} 
We initialize the policy model using Qwen-2.5-14B-Instruct~\citep{qwen}. Our RL framework is implemented with the VeRL~\citep{sheng2024hybridflow} library.

\textbf{Reward.} The reward is computed by \qwenplus~acting as the scorer. Moreover, the scorer is required to explicitly enumerate its reasoning before assigning a score, a step that significantly enhances scoring rigor. Additionally, we observe that the length of the generated idea influences the scorer. 
% Specifically, since we perform comparative scoring, in the early stages of training, extremely short initial ideas tend to yield more pronounced improvements. In later stages, however, the trained research agent tends to generate increasingly longer content to continually secure improvement rewards. 
To mitigate this behavior, we enforce strict length constraints: ideas with fewer than 3,000 or more than 5,000 characters are penalized.

\textbf{Evaluation Protocols and Metrics.} 
% To validate \model, we conducted both automated evaluations and human studies. The human study was extended to three additional domains: law education, and biotechnology.
To validate \model, we conducted both automated evaluation and expert evaluation. The automated evaluation was performed on our constructed validation set, which includes computer science, finance, statistics, and mathematics. The expert evaluation encompassed three new domains: law, education, and biotechnology.

\begin{itemize}[leftmargin=*]

    \item \textbf{Baseline.} We place our \model~alongside the \qwenit~and five leading LLMs, including GPT-4o~\citep{achiam2023gpt}, gemini-2.5-pro~\citep{gemini}, Qwen3-max~\citep{yang2025qwen3technicalreport}, Deepseek-r1~\citep{deepseekr1}, Grok-4.1~\citep{grok4}, and Minimax-M2.1~\citep{Minimax-m2} within the same workflow as shown in Fig.~\ref{Fig:model}. All models receive the contextual input and perform idea generation, with up to three rounds of iterative refinement permitted. The final idea from each model is used for comparison.

\item \textbf{Automated Evaluation.} We begin with automated evaluation. To enhance the reliability of validation, we employ two evaluation approaches: rubrics-based assessment and winrate analysis.  
\noindent\textbf{Rubrics:} We adopt the general rubrics from \citet {goel2025trainingaicoscientistsusing} to evaluate whether the generated idea meet the fundamental rubrics of a scientific research idea. These rubrics are designed based on recurring failure patterns in research ideas generated by language models, as documented in earlier studies~\citep{si2024can}.
% Specifically, we use six general rubrics: (1) Detailed and specific solution, (2) No overlooked fLaws, (3) Well-justified rationale, (4) Cost- and effort-efficient, (5) No ethical issues, and (6) Consistent with the overall plan. 
For each rubric, we determine whether a generated idea satisfies it based on majority voting among the five leading LLMs: GPT-4o~\citep{gpt-4}, Kimi-K2~\citep{kimik2}, gemini-2.5-pro~\citep{gemini}, Deepseek-r1~\citep{deepseekr1}, and Qwen3-max~\citep{yang2025qwen3technicalreport}.  
\noindent\textbf{Winrate:} We compute the average score assigned by the same five LLMs through pairwise comparisons of all generated ideas. We adopt the comparison methodology and prompt template introduced in SGI-bench~\citep{xu2025probing}. SGI-Bench evaluates ideas along four dimensions: \emph{Novelty} (whether the approach is innovative), 
\emph{Feasibility} (whether it can be practically implemented), \emph{Effectiveness} (whether the problem can be solved), and \emph{Detailedness} (whether the proposal is complete and specific), as these four dimensions collectively span the entire pipeline of scientific research.

    % \item \textbf{Expert Evaluation.} To assess the generalization capability of \model, we additionally conduct expert evaluations, extending the evaluation scenarios to three new domains: law, education, and biotechnology. For each domain, we recruit three domain experts. Each evaluator was assigned 10 pairs of ideas, where one idea in each pair was generated by \model~and the other by either GPT-4o or \qwenit. 
    \item \textbf{Expert Evaluation.} To systematically evaluate \model’s adaptability in unseen scenarios and to obtain expert-based validation beyond synthetic benchmarks, we further conducted expert evaluations in three entirely new domains: law, education, and biotechnology. In each domain, we recruited three domain experts. Each evaluator was assigned 10 pairs of research ideas, where one idea in each pair was generated by \model~and the other by either GPT-4o or \qwenit.
    Evaluators were allocated 60 minutes per annotation task and, following the prompt guidelines provided by SGI-Bench, selected a winner for each idea pair across four dimensions: effectiveness, novelty, detailedness, and feasibility. The evaluator can select the ``both bad'' option; once this option is chosen, the pair is excluded from the win rate calculation.

\end{itemize}
For more details of datasets, baselines, and implementation details, please refer to Appendix~\ref{app:exp}.

\subsection{Rubrics Evaluation Result}

\begin{figure*}[t]
    \centering
    \includegraphics[width=0.99\linewidth]{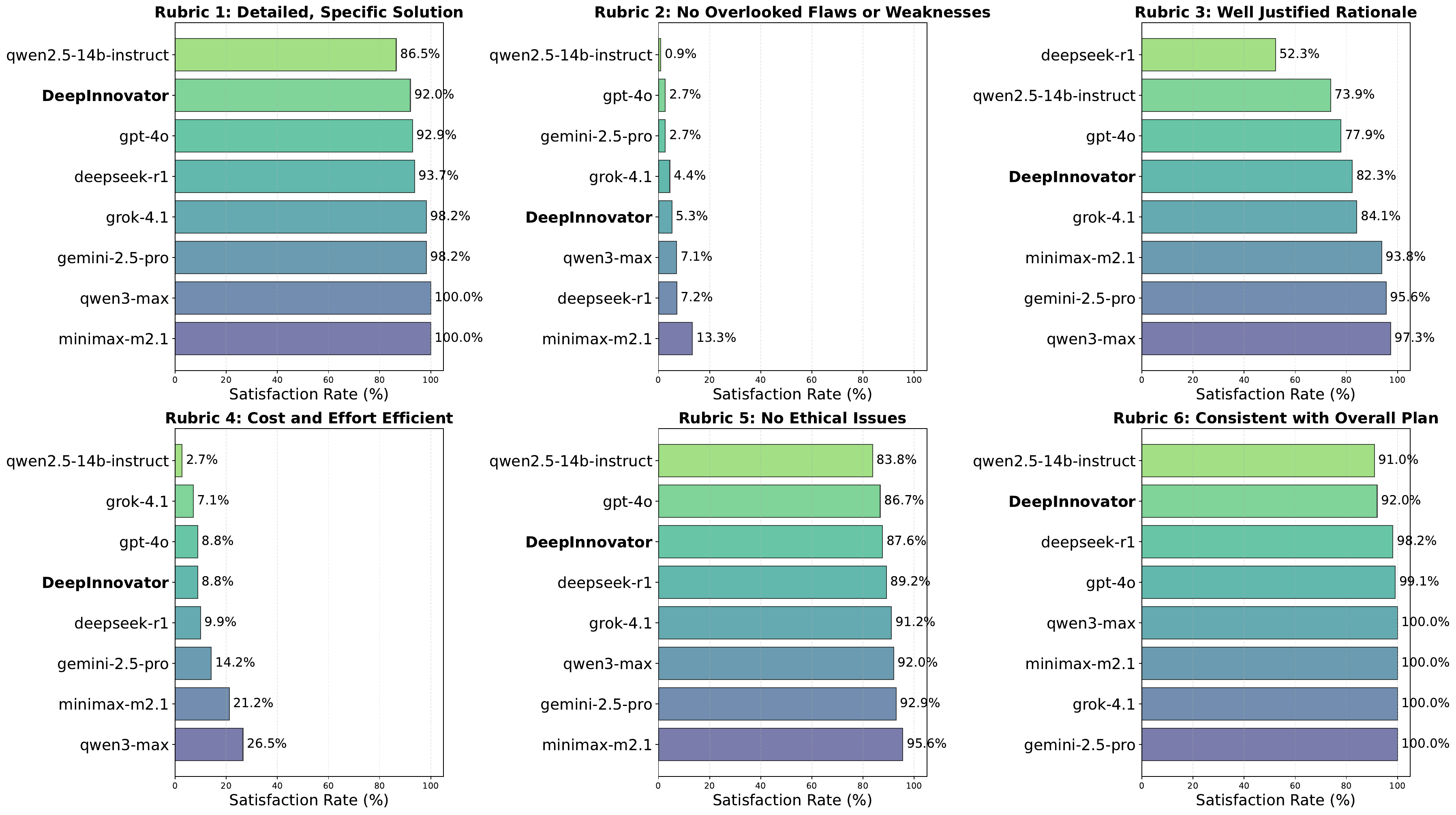}
    \caption{Evaluation results of ideas generated by eight models across six rubrics. \model~outperforms \qwenit~on all dimensions and demonstrates competitive capability against leading LLMs.}
    \label{fig:rubrics result}
\end{figure*}

In the rubrics evaluation, we comprehensively evaluate the solutions generated by models across six core rubrics: 
(1) \emph{Detailed, Specific Solution}, which examines whether the output exhibits actionable steps and depth of detail; 
(2) \emph{No Overlooked Flaws or Weaknesses}, which assesses the model's ability to identify potential issues; 
(3) \emph{Well-Justified Rationale}, which measures the logical coherence and strength of supporting reasoning; 
(4) \emph{Cost and Effort Efficient}, which evaluates the reasonableness of resource utilization in the proposed solution; 
(5) \emph{No Ethical Issues}, which ensures that the recommendations adhere to ethical standards; and 
(6) \emph{Consistent with Overall Plan}, which verifies whether the solution aligns with the overarching contextual objectives.
The details can be found in Sec.~\ref{app:rubrics}.

As shown in Fig.~\ref{fig:rubrics result}, compared to the baseline \qwenit, \model~demonstrates clear performance improvements across multiple dimensions, with gains ranging from 1.05\% to 8.43\%. Furthermore, it approaches the performance (top-5) of top-tier LLMs on 3/6 rubrics. 
Notably, \model~(82.3\%)~surpasses GPT-4o (77.88\%) in \emph{Well-Justified Rationale} (Rubric 3), which emphasizes reasoning detail and argumentation. Additionally, its \emph{No Ethical Issues} (Rubric 5) exceeds that of GPT-4o, reflecting strong safety alignment capabilities.

Moreover, we observe that \emph{No Overlooked Flaws or Weaknesses} (Rubric 2) and \emph{Cost and Effort Efficient} (Rubric 4) present common challenges for current models, reflecting persistent limitations in systematic risk anticipation and resource-optimization modeling. Nevertheless, \model~maintains consistently stable performance across these dimensions.

% \input{table_mainexp}

% \subsection{Performance Analysis (RQ1)}

% \input{table_ablation}
% \subsection{Component-wise Analysis of \model\ (RQ2)}

\subsection{Winrate Evaluation Result}
\begin{table*}[t]
\centering
\caption{Win rate results of \model~vs Other Models.
\emph{Novelty}: whether the approach is innovative; 
\emph{Feasibility}: whether it can be practically implemented; 
\emph{Effectiveness}: whether the problem can be solved;
\emph{Detailedness}: whether the proposal is complete and specific.
\textbf{Bold} indicate that \model~wins (i.e., win rate > 50\%). }
% \vspace{-.5em}
\label{tab:win_ratios_percent}
\resizebox{\textwidth}{!}{%
\begin{tabular}{lccccccc}
\hline
Metric &  {\qwenit} &  {Deepseek-r1} &  {Gemini-2.5-pro} &  {GPT-4o} &  {Minimax-M2.1} &  {Qwen3-max} &  {Grok-4.1} \\
\hline
\emph{Novelty}      & \textbf{80.53\%} & {47.79\%} & 15.93\% & {49.56\%} & \textbf{59.29\%} & 32.74\% & 37.17\% \\
\emph{Feasibility}  & \textbf{87.61\%} & 23.01\% & 18.58\% & 11.50\% & 10.62\% & 19.47\% & 10.62\% \\
\emph{Effectiveness}& \textbf{92.92\%} & {45.13\%} & {43.36\%} & \textbf{76.11\%} & \textbf{50.44\%} & \textbf{66.37\%} & {43.36\%} \\
\emph{Detailedness} & \textbf{93.81\%} & \textbf{59.29\%} & \textbf{68.14\%} & 44.25\% & \textbf{61.06\%} & \textbf{91.15\%} & 45.13\% \\
\hline
\end{tabular}%
}
\end{table*}

Table.~\ref{tab:win_ratios_percent} reports the win rates of \model~on SGI-bench across four key dimensions: \emph{Novelty}, \emph{Effectiveness}, \emph{Feasibility}, and \emph{Detailedness}. All idea pairs were evaluated through majority voting among five independent LLMs to ensure objectivity and consistency in the assessment.

\textbf{The proposed training methodology effectively enhances \model's innovative capability.} As evidenced by the evaluation results, \model~significantly outperforms \qwenit~across all dimensions, demonstrating that our training approach substantially strengthens originality in methodological or conceptual design (\emph{Novelty}) and improves the practical viability of generated ideas (\emph{Feasibility}). Notably, \model~achieves win rates exceeding 90\% against \qwenit~on both  \emph{Effectiveness} and  \emph{Detailedness}, and surpasses the solutions provided by 7/12 larger leading LLMs\footnote{\model~is compared against 6 other LLMs on 2 domains each (\emph{Effectiveness} and \emph{Detailedness}), yielding a total of 12 win rates. ``7/12'' indicates that, among these 12 win rates, 7 exceeded 50\%.}. This indicates that \model~not only efficiently addresses the target tasks but also delivers well-structured and highly detailed solutions.

\textbf{Parameter scale significantly influences innovative capability. }Although \model~outperforms \qwenit, it defeats only 8/24 other LLMs, primarily excelling in the dimensions of  \emph{Effectiveness} and  \emph{Detailedness}. In contrast, it struggles to compete with larger-parameter LLMs in  \emph{Novelty} and  \emph{Feasibility}. Notably, in the feasibility dimension, despite achieving a decisive advantage over \qwenit~(87.61\%), its highest win rate against other models is merely 23.01\%. This highlights that smaller-parameter models face inherent limitations in simultaneously ensuring  \emph{Effectiveness},  \emph{Detailedness}, and  \emph{Feasibility} during innovation.

\begin{figure}[t]
    \centering
    \includegraphics[width=0.6\linewidth]{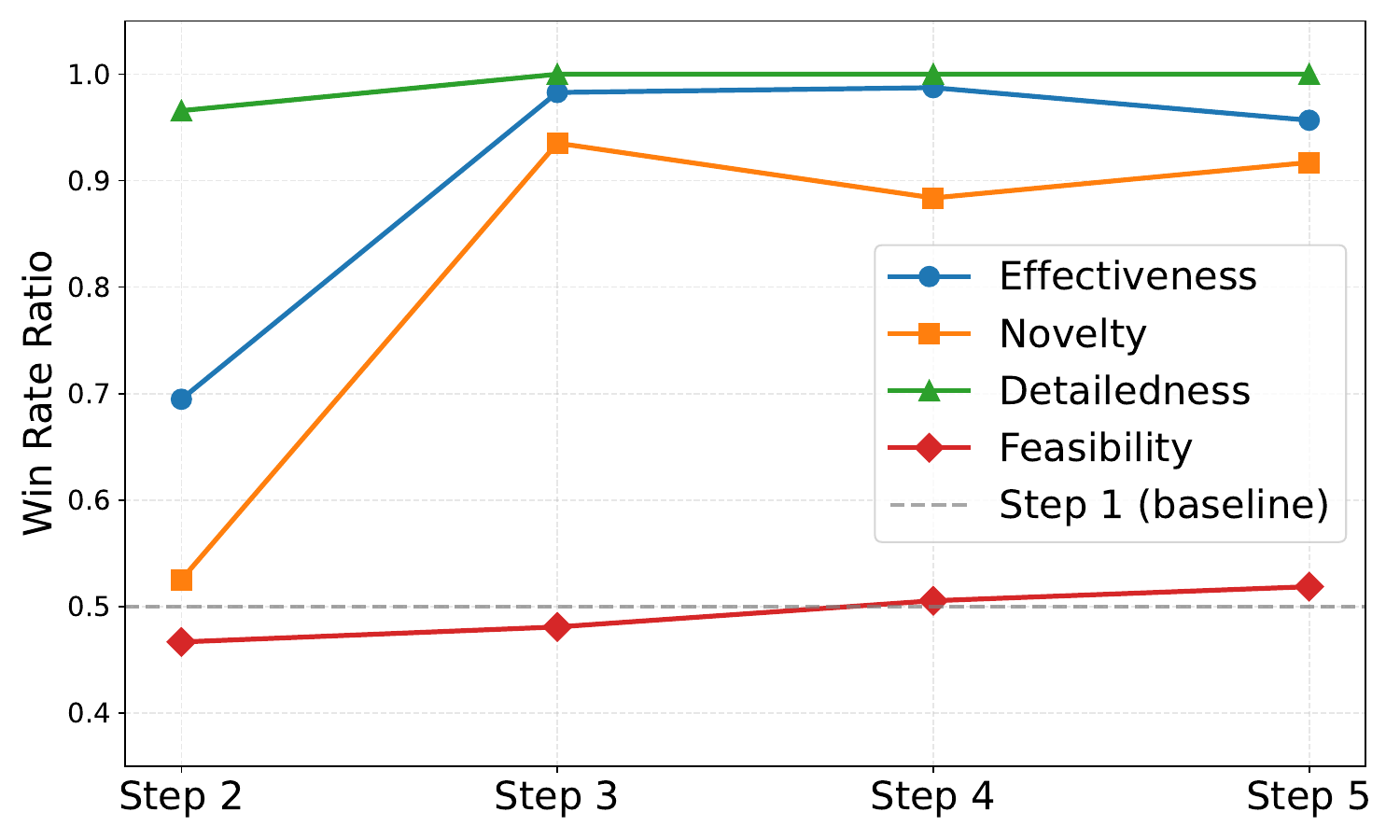}
    \caption{Comparison of win rates between ideas generated by \model~at each step and the initial idea (step 1). Our training method effectively enhances \model's ability to refine research ideas.}
    % \vspace{-2em}
    \label{fig:step_improve}
\end{figure}

\textbf{The refinement process contributes to the improvement of research ideas.}
Fig.~\ref{fig:step_improve} shows the win rates of ideas generated at each step compared to the initial idea (Step 1) across multiple dimensions. Although these evaluation dimensions were not explicitly specified in the training task, the win rates of the newly generated ``next idea'' consistently improve across these dimensions as the refinement process progresses (Steps 2–3), and stabilize during Steps 4–5. This indicates that \model~can effectively enhance the overall quality of ideas during the refinement process.

\begin{table*}[tbp]
\centering
\caption{Winrate results annotated by human experts. ``Much Better'' or ``Much Worse'' indicates that the expert thinks that \model~is much better/worse than \qwenit/GPT-4o. The value before the slash ``/'' denotes \model~vs \qwenit, and the value after the slash denotes \model~vs GPT-4o. The values represent the sum of annotators' judgments. \textbf{Bold} indicate that \model~wins (i.e., win rate > 50\%). }
% \vspace{-.5em}
\label{tab:expert_result}
\resizebox{1.0\textwidth}{!}{%
\begin{tabular}{lccccccc}
\toprule
Domain & Metric & 
Much Better & Better & Worse & Much Worse & Both Bad & Avg. Winrate \\
\midrule

% Domain Law
\multirow{4}{*}{\textbf{Law}}
& \emph{Novelty}        & 0 / 0 & 9 / 7 & 3 / 6 & 1 / 0 & 0 / 0 & \textbf{69.2\%} / \textbf{53.8\%} \\
& \emph{Feasibility}    & 1 / 0 & 5 / 4 & 3 / 7 & 1 / 1 & 3 / 1 & \textbf{60.0\%} / 33.3\% \\
& \emph{Effectiveness}  & 2 / 1 & 5 / 4 & 2 / 3 & 1 / 3 & 3 / 2 & \textbf{70.0\%} / 45.5\% \\
& \emph{Detailedness}   & 7 / 1 & 3 / 4 & 2 / 5 & 1 / 0 & 0 / 3 & \textbf{76.9\%} / 50.0\% \\
\midrule

% Domain Education
\multirow{4}{*}{\textbf{Education}}
& \emph{Novelty}        & 3 / 0 & 9 / 9 & 2 / 5 & 1 / 1 & 0 / 0 & \textbf{80.0\%} / \textbf{60.0\%} \\
& \emph{Feasibility}    & 3 / 0 & 5 / 0 & 5 / 7 & 1 / 3 & 1 / 5 & \textbf{57.1\%} / 0.0\% \\
& \emph{Effectiveness}  & 2 / 0 & 6 / 0 & 4 / 5 & 3 / 4 & 0 / 6 & \textbf{53.3\%} / 0.0\% \\
& \emph{Detailedness}   & 1 / 0 & 8 / 4 & 3 / 5 & 0 / 2 & 3 / 4 & \textbf{75.0\%} / 36.4\% \\
\midrule

% Domain Biotech
\multirow{4}{*}{\textbf{Biotech}}
& \emph{Novelty}        & 3 / 2 & 8 / 6 & 1 / 5 & 0 / 0 & 2 / 2 & \textbf{91.7\%} / \textbf{61.5\%} \\
& \emph{Feasibility}    & 2 / 0 & 6 / 5 & 0 / 3 & 0 / 5 & 6 / 2 & \textbf{100.0\%} / 38.5\% \\
& \emph{Effectiveness}  & 1 / 2 & 6 / 5 & 4 / 2 & 1 / 4 & 2 / 2 & \textbf{58.3\%} / \textbf{53.8\%} \\
& \emph{Detailedness}   & 6 / 3 & 4 / 2 & 1 / 4 & 0 / 5 & 3 / 1 & \textbf{90.9\%} / 35.7\% \\
\bottomrule
\end{tabular}%
}
\end{table*}
\subsection{Expert Evaluation}

Table.~\ref{tab:expert_result} reports the win rates of \model~in three specialized domains outside the training set (law, education, and biotechnology). 
% All idea pairs were evaluated by domain experts. 

\textbf{Our training approach effectively enhances the novelty of \model.} A consistent pattern across all three domains shows that \emph{Novelty} is the only dimension where \model~maintains a win rate above 50\% against both \qwenit~and GPT-4o. This consistent advantage suggests that the training methodology of \model~is particularly effective at fostering original and innovative thinking, regardless of the parameter scale of the baseline models. Even when compared to large models such as GPT-4o, \model~remains competitive in terms of \emph{Novelty}.

\textbf{The generalizability of innovative capabilities is challenged.} The performance of \model~still exhibits domain dependence. Although \model~achieves consistently superior results against \qwenit~across all domains, its performance in the education domain is relatively weaker—particularly when compared to GPT-4o, where win rates for both \emph{Feasibility} and \emph{Effectiveness} drop to 0.0\%. In the legal domain, \model~demonstrates balanced performance: all dimensions exceed 50\% win rates against \qwenit, yet only \emph{Novelty} maintains an advantage over GPT-4o (53.8\%). This domain-specific variation suggests that the training data or methodology of \model~is better aligned with STEM-oriented fields such as biotechnology, while offering insufficient advantages in humanities-related domains.

\textbf{The feasibility dimension exhibits the largest performance gap in baseline comparisons.} In the biotechnology domain, \model~received no judgments of ``Worse'' or ``Much Worse'' when compared against \qwenit~on \emph{Feasibility}. However, comparisons with GPT-4o reveal significant challenges: win rates drop to 33.3\% (law), 0.0\% (education), and 38.5\% (biotechnology). Notably, in the education domain, \model~ failed to win a single feasibility comparison (0.0\% win rate). This substantial gap suggests that larger models possess stronger capabilities in evaluating and generating practically feasible solutions, indicating that improving feasibility may require broader world knowledge and more advanced reasoning abilities.

% \input{sec_eval_casestudy}
% \input{relate}%放到附录了
% \vspace{-1em}
\section{Conclusion}
We explores how to stimulate the innovative capabilities of LLMs, specifically manifested in the core activities of autonomous scientific research: the generation and refinement of research ideas. We propose the \model~framework, which (1) automatically extracts and synthesizes structured knowledge from vast scientific literature, providing the model with a solid \emph{``shoulders of giants''}; and (2) employs a ``Next Idea Prediction'' training paradigm, constructing a process reward and a decoupled reward-comment mechanism that effectively simulates the cycle of \emph{``conjectures and refutations''}—a self-critical, iterative improvement process inherent in scientific inquiry.
Experiments demonstrate that \model-14B generates high-quality, novel research ideas and exhibits exceptional cross-domain generalization. Even in domains absent from its training data (law, education, and biotechnology), \model-14B consistently outperforms \qwenit~ and rivals leading LLMs. This result indicates that a carefully designed training framework can effectively stimulate the innovative capability of LLMs. 
% Our work lays a crucial foundation for building research agents capable of genuinely accelerating scientific discovery.
Our work explores important training methodologies for scientific discovery AI, effectively triggers the innovative capabilities of research agents.

% \clearpage
% \section*{Impact Statement}
% This paper presents work whose goal is to advance the field of 
% Machine Learning. There are many potential societal consequences 
% of our work, none which we feel must be specifically highlighted here.


\clearpage
\bibliographystyle{iclr}
\bibliography{refs}

@article{achiam2023gpt,
  title={Gpt-4 technical report},
  author={Achiam, Josh and Adler, Steven and Agarwal, Sandhini and Ahmad, Lama and Akkaya, Ilge and Aleman, Florencia Leoni and Almeida, Diogo and Altenschmidt, Janko and Altman, Sam and Anadkat, Shyamal and others},
  journal={arXiv preprint arXiv:2303.08774},
  year={2023}
}

@misc{grok4,
  author={{xAI}},
  title={Grok-4.1},
  year={2025},
  howpublished={\url{https://data.x.ai/2025-11-17-grok-4-1-model-card.pdf}}
}

@misc{Minimax-m2,
  author={{MiniMax}},
  title={MiniMax-M2.1},
  year={2025},
  howpublished={\url{https://github.com/MiniMax-AI/MiniMax-M2.1}}
}

@article{yamada2025ai,
  title={The ai scientist-v2: Workshop-level automated scientific discovery via agentic tree search},
  author={Yamada, Yutaro and Lange, Robert Tjarko and Lu, Cong and Hu, Shengran and Lu, Chris and Foerster, Jakob and Clune, Jeff and Ha, David},
  journal={arXiv preprint arXiv:2504.08066},
  year={2025}
}

@article{Gao2023ScalingLF,
  title={Scaling Laws for Reward Model Overoptimization},
  author={Gao, Long and Schulman, John and Hilton, Jacob},
  journal={Proceedings of the 40th International Conference on Machine Learning (ICML)},
  year={2023},
  volume={202},
  pages={10666-10682},
  url={https://proceedings.mlr.press/v202/gao23h.html}
}

@article{Zhao2025OneTT,
  title={One Token to Fool LLM-as-a-Judge},
  author={Zhao, Yulai and Liu, Haolin and Yu, Dian and Kung, Sunyuan and Chen, Meijia and Mi, Haitao and Yu, Dong},
  journal={arXiv preprint arXiv:2507.08794},
  year={2025}
}

@misc{goel2025trainingaicoscientistsusing,
      title={Training AI Co-Scientists Using Rubric Rewards}, 
      author={Shashwat Goel and Rishi Hazra and Dulhan Jayalath and Timon Willi and Parag Jain and William F. Shen and Ilias Leontiadis and Francesco Barbieri and Yoram Bachrach and Jonas Geiping and Chenxi Whitehouse},
      year={2025},
      eprint={2512.23707},
      archivePrefix={arXiv},
      primaryClass={cs.LG},
      url={https://arxiv.org/abs/2512.23707}, 
}

@article{kimik2,
  title={Kimi k2: Open agentic intelligence},
  author={Team, Kimi and Bai, Yifan and Bao, Yiping and Chen, Guanduo and Chen, Jiahao and Chen, Ningxin and Chen, Ruijue and Chen, Yanru and Chen, Yuankun and Chen, Yutian and others},
  journal={arXiv preprint arXiv:2507.20534},
  year={2025}
}

@article{deepseekr1,
   title={DeepSeek-R1 incentivizes reasoning in LLMs through reinforcement learning},
   volume={645},
   ISSN={1476-4687},
   url={http://dx.doi.org/10.1038/s41586-025-09422-z},
   DOI={10.1038/s41586-025-09422-z},
   number={8081},
   journal={Nature},
   publisher={Springer Science and Business Media LLC},
   author={Guo, Daya and Yang, Dejian and Zhang, Haowei and Song, Junxiao and Wang, Peiyi and Zhu, Qihao and Xu, Runxin and Zhang, Ruoyu and Ma, Shirong and Bi, Xiao and Zhang, Xiaokang and Yu, Xingkai and Wu, Yu and Wu, Z. F. and Gou, Zhibin and Shao, Zhihong and Li, Zhuoshu and Gao, Ziyi and Liu, Aixin and Xue, Bing and Wang, Bingxuan and Wu, Bochao and Feng, Bei and Lu, Chengda and Zhao, Chenggang and Deng, Chengqi and Ruan, Chong and Dai, Damai and Chen, Deli and Ji, Dongjie and Li, Erhang and Lin, Fangyun and Dai, Fucong and Luo, Fuli and Hao, Guangbo and Chen, Guanting and Li, Guowei and Zhang, H. and Xu, Hanwei and Ding, Honghui and Gao, Huazuo and Qu, Hui and Li, Hui and Guo, Jianzhong and Li, Jiashi and Chen, Jingchang and Yuan, Jingyang and Tu, Jinhao and Qiu, Junjie and Li, Junlong and Cai, J. L. and Ni, Jiaqi and Liang, Jian and Chen, Jin and Dong, Kai and Hu, Kai and You, Kaichao and Gao, Kaige and Guan, Kang and Huang, Kexin and Yu, Kuai and Wang, Lean and Zhang, Lecong and Zhao, Liang and Wang, Litong and Zhang, Liyue and Xu, Lei and Xia, Leyi and Zhang, Mingchuan and Zhang, Minghua and Tang, Minghui and Zhou, Mingxu and Li, Meng and Wang, Miaojun and Li, Mingming and Tian, Ning and Huang, Panpan and Zhang, Peng and Wang, Qiancheng and Chen, Qinyu and Du, Qiushi and Ge, Ruiqi and Zhang, Ruisong and Pan, Ruizhe and Wang, Runji and Chen, R. J. and Jin, R. L. and Chen, Ruyi and Lu, Shanghao and Zhou, Shangyan and Chen, Shanhuang and Ye, Shengfeng and Wang, Shiyu and Yu, Shuiping and Zhou, Shunfeng and Pan, Shuting and Li, S. S. and Zhou, Shuang and Wu, Shaoqing and Yun, Tao and Pei, Tian and Sun, Tianyu and Wang, T. and Zeng, Wangding and Liu, Wen and Liang, Wenfeng and Gao, Wenjun and Yu, Wenqin and Zhang, Wentao and Xiao, W. L. and An, Wei and Liu, Xiaodong and Wang, Xiaohan and Chen, Xiaokang and Nie, Xiaotao and Cheng, Xin and Liu, Xin and Xie, Xin and Liu, Xingchao and Yang, Xinyu and Li, Xinyuan and Su, Xuecheng and Lin, Xuheng and Li, X. Q. and Jin, Xiangyue and Shen, Xiaojin and Chen, Xiaosha and Sun, Xiaowen and Wang, Xiaoxiang and Song, Xinnan and Zhou, Xinyi and Wang, Xianzu and Shan, Xinxia and Li, Y. K. and Wang, Y. Q. and Wei, Y. X. and Zhang, Yang and Xu, Yanhong and Li, Yao and Zhao, Yao and Sun, Yaofeng and Wang, Yaohui and Yu, Yi and Zhang, Yichao and Shi, Yifan and Xiong, Yiliang and He, Ying and Piao, Yishi and Wang, Yisong and Tan, Yixuan and Ma, Yiyang and Liu, Yiyuan and Guo, Yongqiang and Ou, Yuan and Wang, Yuduan and Gong, Yue and Zou, Yuheng and He, Yujia and Xiong, Yunfan and Luo, Yuxiang and You, Yuxiang and Liu, Yuxuan and Zhou, Yuyang and Zhu, Y. X. and Huang, Yanping and Li, Yaohui and Zheng, Yi and Zhu, Yuchen and Ma, Yunxian and Tang, Ying and Zha, Yukun and Yan, Yuting and Ren, Z. Z. and Ren, Zehui and Sha, Zhangli and Fu, Zhe and Xu, Zhean and Xie, Zhenda and Zhang, Zhengyan and Hao, Zhewen and Ma, Zhicheng and Yan, Zhigang and Wu, Zhiyu and Gu, Zihui and Zhu, Zijia and Liu, Zijun and Li, Zilin and Xie, Ziwei and Song, Ziyang and Pan, Zizheng and Huang, Zhen and Xu, Zhipeng and Zhang, Zhongyu and Zhang, Zhen},
   year={2025},
   month=sep, pages={633–638} }

@misc{yang2025qwen3technicalreport,
      title={Qwen3 Technical Report}, 
      author={An Yang and Anfeng Li and Baosong Yang and Beichen Zhang and Binyuan Hui and Bo Zheng and Bowen Yu and Chang Gao and Chengen Huang and Chenxu Lv and Chujie Zheng and Dayiheng Liu and Fan Zhou and Fei Huang and Feng Hu and Hao Ge and Haoran Wei and Huan Lin and Jialong Tang and Jian Yang and Jianhong Tu and Jianwei Zhang and Jianxin Yang and Jiaxi Yang and Jing Zhou and Jingren Zhou and Junyang Lin and Kai Dang and Keqin Bao and Kexin Yang and Le Yu and Lianghao Deng and Mei Li and Mingfeng Xue and Mingze Li and Pei Zhang and Peng Wang and Qin Zhu and Rui Men and Ruize Gao and Shixuan Liu and Shuang Luo and Tianhao Li and Tianyi Tang and Wenbiao Yin and Xingzhang Ren and Xinyu Wang and Xinyu Zhang and Xuancheng Ren and Yang Fan and Yang Su and Yichang Zhang and Yinger Zhang and Yu Wan and Yuqiong Liu and Zekun Wang and Zeyu Cui and Zhenru Zhang and Zhipeng Zhou and Zihan Qiu},
      year={2025},
      eprint={2505.09388},
      archivePrefix={arXiv},
      primaryClass={cs.CL},
      url={https://arxiv.org/abs/2505.09388}, 
}

@inproceedings{collabllm2025,
    title={CollabLLM: From Passive Responders to Active Collaborators},
    author={Shirley Wu and Michel Galley and Baolin Peng and Hao Cheng and 
            Gavin Li and Yao Dou and Weixin Cai and James Zou and 
            Jure Leskovec and Jianfeng Gao},
    booktitle={International Conference on Machine Learning (ICML)},
    year={2025}
}

@article{xu2025probing,
  title={Probing Scientific General Intelligence of LLMs with Scientist-Aligned Workflows},
  author={Xu, Wanghan and Zhou, Yuhao and Zhou, Yifan and Cao, Qinglong and Li, Shuo and Bu, Jia and Liu, Bo and Chen, Yixin and He, Xuming and Zhao, Xiangyu and others},
  journal={arXiv preprint arXiv:2512.16969},
  year={2025}
}

@article{sheng2024hybridflow,
  title   = {HybridFlow: A Flexible and Efficient RLHF Framework},
  author  = {Guangming Sheng and Chi Zhang and Zilingfeng Ye and Xibin Wu and Wang Zhang and Ru Zhang and Yanghua Peng and Haibin Lin and Chuan Wu},
  year    = {2024},
  journal = {arXiv preprint arXiv: 2409.19256}
}

@article{qwen, 
title={Qwen Technical Report}, 
author={Jinze Bai and Shuai Bai and Yunfei Chu and Zeyu Cui and Kai Dang and Xiaodong Deng and Yang Fan and Wenbin Ge and Yu Han and Fei Huang and Binyuan Hui and Luo Ji and Mei Li and Junyang Lin and Runji Lin and Dayiheng Liu and Gao Liu and Chengqiang Lu and Keming Lu and Jianxin Ma and Rui Men and Xingzhang Ren and Xuancheng Ren and Chuanqi Tan and Sinan Tan and Jianhong Tu and Peng Wang and Shijie Wang and Wei Wang and Shengguang Wu and Benfeng Xu and Jin Xu and An Yang and Hao Yang and Jian Yang and Shusheng Yang and Yang Yao and Bowen Yu and Hongyi Yuan and Zheng Yuan and Jianwei Zhang and Xingxuan Zhang and Yichang Zhang and Zhenru Zhang and Chang Zhou and Jingren Zhou and Xiaohuan Zhou and Tianhang Zhu}, 
journal={arXiv preprint arXiv:2309.16609}, 
year={2023} }

@article{Hu2025BiasFT,
  title={Bias Fitting to Mitigate Length Bias of Reward Model in RLHF},
  author={Hu, Yutong and Ouyang, Shuai and Li, Qinyuan and Yi, Hao and Chen, Guodong and Zhang, Fan and Li, Xiaofei},
  journal={arXiv preprint arXiv:2505.12843},
  year={2025}
}

@misc{Anthropic2025NaturalEM,
  title={Natural emergent misalignment from reward hacking},
  author={Anthropic},
  howpublished={Anthropic Research Blog},
  year={2025},
  url={https://www.anthropic.com/research/emergent-misalignment-reward-hacking}
}

@article{shao2024deepseekmath,
  title={Deepseekmath: Pushing the limits of mathematical reasoning in open language models},
  author={Shao, Zhihong and Wang, Peiyi and Zhu, Qihao and Xu, Runxin and Song, Junxiao and Bi, Xiao and Zhang, Haowei and Zhang, Mingchuan and Li, YK and Wu, Yang and others},
  journal={arXiv preprint arXiv:2402.03300},
  year={2024}
}

@article{shao2025dr,
  title={Dr tulu: Reinforcement learning with evolving rubrics for deep research},
  author={Shao, Rulin and Asai, Akari and Shen, Shannon Zejiang and Ivison, Hamish and Kishore, Varsha and Zhuo, Jingming and Zhao, Xinran and Park, Molly and Finlayson, Samuel G and Sontag, David and others},
  journal={arXiv preprint arXiv:2511.19399},
  year={2025}
}

@article{seed2025seed1,
  title={Seed1. 5-thinking: Advancing superb reasoning models with reinforcement learning},
  author={Seed, ByteDance and Chen, Jiaze and Fan, Tiantian and Liu, Xin and Liu, Lingjun and Lin, Zhiqi and Wang, Mingxuan and Wang, Chengyi and Wei, Xiangpeng and Xu, Wenyuan and others},
  journal={arXiv preprint arXiv:2504.13914},
  year={2025}
}

@article{wang2025harnessing,
  title={Harnessing uncertainty: Entropy-modulated policy gradients for long-horizon llm agents},
  author={Wang, Jiawei and Liu, Jiacai and Fu, Yuqian and Li, Yingru and Wang, Xintao and Lin, Yuan and Yue, Yu and Zhang, Lin and Wang, Yang and Wang, Ke},
  journal={arXiv preprint arXiv:2509.09265},
  year={2025}
}

@incollection{gentner2011analogy,
  title={Analogy in scientific discovery: The case of Johannes Kepler},
  author={Gentner, Dedre},
  booktitle={Model-based reasoning: Science, technology, values},
  pages={21--39},
  year={2011},
  publisher={Springer}
}

@article{kang2022augmenting,
  title={Augmenting scientific creativity with retrieval across knowledge domains},
  author={Kang, Hyeonsu B and Mysore, Sheshera and Huang, Kevin and Chang, Haw-Shiuan and Prein, Thorben and McCallum, Andrew and Kittur, Aniket and Olivetti, Elsa},
  journal={arXiv preprint arXiv:2206.01328},
  year={2022}
}

@article{kennedy2022serendipity,
  title={Serendipity: A way of stimulating researchers' creativity},
  author={Kennedy, Ian G and Whitehead, Dean and Ferdinand-James, Debra},
  journal={Journal of Creativity},
  volume={32},
  number={1},
  pages={100014},
  year={2022},
  publisher={Elsevier}
}

@article{min2021identifying,
  title={Identifying citation patterns of scientific breakthroughs: A perspective of dynamic citation process},
  author={Min, Chao and Bu, Yi and Wu, Ding and Ding, Ying and Zhang, Yi},
  journal={Information Processing \& Management},
  volume={58},
  number={1},
  pages={102428},
  year={2021},
  publisher={Elsevier}
}

@article{chen2006citespace,
  title={CiteSpace II: Detecting and visualizing emerging trends and transient patterns in scientific literature},
  author={Chen, Chaomei},
  journal={Journal of the American Society for information Science and Technology},
  volume={57},
  number={3},
  pages={359--377},
  year={2006},
  publisher={Wiley Online Library}
}

@article{harvey2014creative,
  title={Creative synthesis: Exploring the process of extraordinary group creativity},
  author={Harvey, Sarah},
  journal={Academy of management review},
  volume={39},
  number={3},
  pages={324--343},
  year={2014},
  publisher={Academy of Management Briarcliff Manor, NY}
}

@article{newell2001theory,
  title={A theory of interdisciplinary studies},
  author={Newell, William H and Wentworth, Jay and Sebberson, David},
  journal={Issues in Interdisciplinary Studies},
  year={2001},
  publisher={Association for Interdisciplinary Studies}
}

@article{weng2025deepscientist,
  title={Deepscientist: Advancing frontier-pushing scientific findings progressively},
  author={Weng, Yixuan and Zhu, Minjun and Xie, Qiujie and Sun, Qiyao and Lin, Zhen and Liu, Sifan and Zhang, Yue},
  journal={arXiv preprint arXiv:2509.26603},
  year={2025}
}

@article{si2024can,
  title={Can llms generate novel research ideas? a large-scale human study with 100+ nlp researchers},
  author={Si, Chenglei and Yang, Diyi and Hashimoto, Tatsunori},
  journal={arXiv preprint arXiv:2409.04109},
  year={2024}
}

@article{tang2025ai,
  title={AI-Researcher: Autonomous Scientific Innovation},
  author={Tang, Jiabin and Xia, Lianghao and Li, Zhonghang and Huang, Chao},
  journal={arXiv preprint arXiv:2505.18705},
  year={2025}
}

@article{jumper2021highly,
  title={Highly accurate protein structure prediction with AlphaFold},
  author={Jumper, John and Evans, Richard and Pritzel, Alexander and Green, Tim and Figurnov, Michael and Ronneberger, Olaf and Tunyasuvunakool, Kathryn and Bates, Russ and {\v{Z}}{\'\i}dek, Augustin and Potapenko, Anna and others},
  journal={nature},
  volume={596},
  number={7873},
  pages={583--589},
  year={2021},
  publisher={Nature Publishing Group UK London}
}

@article{novikov2025alphaevolve,
  title={AlphaEvolve: A coding agent for scientific and algorithmic discovery},
  author={Novikov, Alexander and V{\~u}, Ng{\^a}n and Eisenberger, Marvin and Dupont, Emilien and Huang, Po-Sen and Wagner, Adam Zsolt and Shirobokov, Sergey and Kozlovskii, Borislav and Ruiz, Francisco JR and Mehrabian, Abbas and others},
  journal={arXiv preprint arXiv:2506.13131},
  year={2025}
}

@String(ICML = {International Conference on Machine Learning})

@misc{gpt-4,
  title       = "GPT-4 Technical Report",
  author      = "OpenAI",
  howpublished = "\url{https://cdn.openai.com/papers/gpt-4.pdf}",
  year        = {2023},
}

@misc{gemini,
  title       = {Gemini: A Family of Highly Capable Multimodal Models},
  author      = {Google},
  howpublished = "\url{https://goo.gle/GeminiPaper}",
  year        = {2023},
}

@article{xu2025comprehensive,
  title={A Comprehensive Survey of Deep Research: Systems, Methodologies, and Applications},
  author={Xu, Renjun and Peng, Jingwen},
  journal={arXiv preprint arXiv:2506.12594},
  year={2025}
}

@article{zhang2025deep,
  title={Deep research: A survey of autonomous research agents},
  author={Zhang, Wenlin and Li, Xiaopeng and Zhang, Yingyi and Jia, Pengyue and Wang, Yichao and Guo, Huifeng and Liu, Yong and Zhao, Xiangyu},
  journal={arXiv preprint arXiv:2508.12752},
  year={2025}
}

@misc{openaidr,
    title = {Introducing deep research},
    url = {https://openai.com/index/introducing-deep-research/},
    author = {OpenAI},
    month = {February},
    year = {2025}
}

@article{fan2025understanding,
  title={Understanding DeepResearch via Reports},
  author={Fan, Tianyu and Niu, Xinyao and Zheng, Yuxiang and Zhang, Fengji and Huang, Chengen and Chen, Bei and Lin, Junyang and Huang, Chao},
  journal={arXiv preprint arXiv:2510.07861},
  year={2025}
}

@article{jin2025search,
  title={Search-r1: Training llms to reason and leverage search engines with reinforcement learning},
  author={Jin, Bowen and Zeng, Hansi and Yue, Zhenrui and Yoon, Jinsung and Arik, Sercan and Wang, Dong and Zamani, Hamed and Han, Jiawei},
  journal={arXiv preprint arXiv:2503.09516},
  year={2025}
}

@article{fan2025posterior,
  title={Posterior-grpo: Rewarding reasoning processes in code generation},
  author={Fan, Lishui and Zhang, Yu and Chen, Mouxiang and Liu, Zhongxin},
  journal={arXiv preprint arXiv:2508.05170},
  year={2025}
}

@article{viswanathan2025checklists,
  title={Checklists are better than reward models for aligning language models},
  author={Viswanathan, Vijay and Sun, Yanchao and Ma, Shuang and Kong, Xiang and Cao, Meng and Neubig, Graham and Wu, Tongshuang},
  journal={arXiv preprint arXiv:2507.18624},
  year={2025}
}

@article{gunjal2025rubrics,
  title={Rubrics as rewards: Reinforcement learning beyond verifiable domains},
  author={Gunjal, Anisha and Wang, Anthony and Lau, Elaine and Nath, Vaskar and He, Yunzhong and Liu, Bing and Hendryx, Sean},
  journal={arXiv preprint arXiv:2507.17746},
  year={2025}
}

@inproceedings{lee2024rlaif,
  title={RLAIF vs. RLHF: scaling reinforcement learning from human feedback with AI feedback},
  author={Lee, Harrison and Phatale, Samrat and Mansoor, Hassan and Mesnard, Thomas and Ferret, Johan and Lu, Kellie and Bishop, Colton and Hall, Ethan and Carbune, Victor and Rastogi, Abhinav and others},
  booktitle={Proceedings of the 41st International Conference on Machine Learning},
  pages={26874--26901},
  year={2024}
}

@article{sharma2024critical,
  title={A critical evaluation of ai feedback for aligning large language models},
  author={Sharma, Archit and Keh, Sedrick Scott and Mitchell, Eric and Finn, Chelsea and Arora, Kushal and Kollar, Thomas},
  journal={Advances in Neural Information Processing Systems},
  volume={37},
  pages={29166--29190},
  year={2024}
}

@inproceedings{maeureka,
  title={Eureka: Human-Level Reward Design via Coding Large Language Models},
  author={Ma, Yecheng Jason and Liang, William and Wang, Guanzhi and Huang, De-An and Bastani, Osbert and Jayaraman, Dinesh and Zhu, Yuke and Fan, Linxi and Anandkumar, Anima},
  booktitle={The Twelfth International Conference on Learning Representations}
}

@article{huang2025reinforcement,
  title={Reinforcement learning with rubric anchors},
  author={Huang, Zenan and Zhuang, Yihong and Lu, Guoshan and Qin, Zeyu and Xu, Haokai and Zhao, Tianyu and Peng, Ru and Hu, Jiaqi and Shen, Zhanming and Hu, Xiaomeng and others},
  journal={arXiv preprint arXiv:2508.12790},
  year={2025}
}

@article{merchant2023scaling,
  title={Scaling deep learning for materials discovery},
  author={Merchant, Amil and Batzner, Simon and Schoenholz, Samuel S and Aykol, Muratahan and Cheon, Gowoon and Cubuk, Ekin Dogus},
  journal={Nature},
  volume={624},
  number={7990},
  pages={80--85},
  year={2023},
  publisher={Nature Publishing Group UK London}
}

@article{newtoncorrespondence,
  title={The Correspondence of Isaac Newton},
  author={Newton, Isaac}
}

@book{popper2014conjectures,
  title={Conjectures and refutations: The growth of scientific knowledge},
  author={Popper, Karl},
  year={2014},
  publisher={routledge}
}

@article{zheng2025deepresearcher,
  title={Deepresearcher: Scaling deep research via reinforcement learning in real-world environments},
  author={Zheng, Yuxiang and Fu, Dayuan and Hu, Xiangkun and Cai, Xiaojie and Ye, Lyumanshan and Lu, Pengrui and Liu, Pengfei},
  journal={arXiv preprint arXiv:2504.03160},
  year={2025}
}

%%%%%%%%%%%%%%%%%%%%%%%%%%%%%%%%%%%%%%%%%%%%%%%%%%%%%%%%%%%%%%%%%%%%%%%%%%%%%%%
%%%%%%%%%%%%%%%%%%%%%%%%%%%%%%%%%%%%%%%%%%%%%%%%%%%%%%%%%%%%%%%%%%%%%%%%%%%%%%%
% APPENDIX
%%%%%%%%%%%%%%%%%%%%%%%%%%%%%%%%%%%%%%%%%%%%%%%%%%%%%%%%%%%%%%%%%%%%%%%%%%%%%%%
%%%%%%%%%%%%%%%%%%%%%%%%%%%%%%%%%%%%%%%%%%%%%%%%%%%%%%%%%%%%%%%%%%%%%%%%%%%%%%%
\newpage
\appendix
\onecolumn
\appendix
\onecolumn

\renewcommand\thefigure{A\arabic{figure}}
\renewcommand\thetable{A\arabic{table}}
\renewcommand\theequation{A.\arabic{equation}}
\renewcommand\thetheorem{A.\arabic{theorem}}
\setcounter{table}{0}
\setcounter{figure}{0}
\setcounter{theorem}{0}
\setcounter{equation}{0}

\begin{center}
\huge {\textbf{Appendix}}    
\end{center}

\normalsize
\section{Experiment Details.}
\label{app:exp}
\subsection{Datasets.}
We employ the official arXiv API\footnote{https://info.arxiv.org/help/api/index.html} to crawl papers. We use Qwen-OCR\footnote{https://www.alibabacloud.com/help/en/model-studio/qwen-vl-ocr} for PDF parsing and Qwen3-max~\citep{yang2025qwen3technicalreport} to extract references from the parsed paper text.  Both our training and validation datasets were partitioned into four disciplinary categories: Computer Science (CS), Mathematics (math), Quantitative Finance (q-fin), and Statistics (stat). Our training set comprises 1,012 samples, distributed as follows: cs: 433, math: 194, q-fin: 181, stat: 204. Our validation set contains 113 samples, with the following distribution: cs: 51, math: 22, q-fin: 20, stat: 20.

\subsection{Implementation Details.} 
The training process utilized multi-turn dialogue data and was optimized with a custom dialogue-level reward function. Regarding the training script, we referred to the implementation of CollabLLM~\citep{collabllm2025}.

The key hyperparameter settings are listed in the following table:

\begin{table}[h]
\centering
\caption{Hyperparameters used in \model.}
\label{tab:hyperparams}
\begin{tabular}{lll}
\toprule
\textbf{Category} & \textbf{Parameter} & \textbf{Value} \\
\midrule
\multicolumn{3}{l}{\textit{Dataset Statistics}} \\
& Training set size (by category) & cs: 433,\, math: 194,\, q-fin: 181,\, stat: 204 \\
& Validation set size (by category) & cs: 51,\, math: 22,\, q-fin: 18,\, stat: 20 \\
\midrule
\multicolumn{3}{l}{\textit{Training Data}} \\
& Train batch size & 16 \\
& Max prompt length & 8192 tokens \\
& Max response length & 2048 tokens \\
& Filter overlong prompts & True \\
\midrule
\multicolumn{3}{l}{\textit{PPO Algorithm}} \\
& PPO mini-batch size & 8 \\
& Advantage estimator & GRPO \\
& KL loss coefficient & 0.001 \\
& KL loss type & low\_var\_kl \\
\midrule
\multicolumn{3}{l}{\textit{Optimizer}} \\
& Actor learning rate & $5 \times 10^{-7}$ \\
& Critic warmup epochs & 0 \\
\midrule
\multicolumn{3}{l}{\textit{Rollout Configuration}} \\
& Number of rollouts & 8 \\
& Temperature & 1.0 \\
& Max user turns & 5 \\
& Max assistant turns & 6 \\
& Repeat rollouts per trajectory & 3 \\
\bottomrule
\end{tabular}
\end{table}

\textbf{Reward.} The reward is computed by \qwenplus~acting as the scorer. Moreover, the scorer is required to explicitly enumerate its reasoning before assigning a score, a step that significantly enhances scoring rigor. Additionally, we observe that the length of the generated idea influences the scorer. Specifically, since we perform comparative scoring, in the early stages of training, extremely short initial ideas tend to yield more pronounced improvements. In later stages, however, the trained research agent tends to generate increasingly longer content to continually secure improvement rewards. To mitigate this behavior, we enforce strict length constraints: ideas with fewer than 3,000 or more than 5,000 characters are penalized.

\subsection{Rubrics Evaluation Details}
\label{app:rubrics}
We employ 6 basic rubrics from~\citet{goel2025trainingaicoscientistsusing} for evaluation. The specific details of these assessments are presented in Table~\ref{apptab:rubrics}.

\begin{table}[ht]
\centering
\caption{Evaluation Rubrics.}
\label{apptab:rubrics}
\begin{tabular}{@{}>{\bfseries}l p{7cm}@{}}
\toprule
\textbf{Rubric} & \textbf{Description} \\
\midrule
Detailed, specific solution &
Does the part of the plan relevant to satisfying this rubric item include fully specified details on how to implement it? There should be no claims of handling something without actually doing so, no vague terms, ambiguity, or lack of clarity. The description should use simple, easy-to-understand language. \\
\addlinespace

No overlooked flaws or weaknesses &
Are there any important overlooked flaws or weaknesses in the part of the plan addressing this rubric item that would invalidate its claimed satisfaction of the criterion? \\
\addlinespace

Well-justified rationale &
Is the part of the plan relevant to this rubric item well-motivated and justified? For example, does it provide convincing arguments that the chosen approach is better than simpler alternatives or competing hypotheses? \\
\addlinespace

Cost and effort efficient &
Does the plan address this rubric item in a cost- and effort-efficient manner, avoiding unnecessary complexity? Consider whether a less resource-intensive or less labor-demanding solution could achieve comparable effectiveness. \\
\addlinespace

No ethical issues &
Does this part of the plan pose any potential for negative societal impact or raise ethical concerns (e.g., bias, privacy violations, manipulation, or unfair outcomes)? \\
\addlinespace

Consistent with overall plan &
Is this component of the plan logically consistent with the rest of the proposed strategy? Ensure it does not contradict other stated assumptions, methods, or objectives elsewhere in the plan. \\
\bottomrule
\end{tabular}
\end{table}
\subsection{Winrate Evaluation Details}

SGI-Bench~\citep{xu2025probing} evaluates ideas along four dimensions: \textbf{effectiveness} (whether the problem can be solved), \textbf{novelty} (whether the approach is innovative), \textbf{detailedness} (whether the proposal is complete and specific), and \textbf{feasibility} (whether it can be practically implemented), as these four dimensions collectively span the entire pipeline of scientific research.

In the original idea evaluation framework of SGI-Bench, the final score for each dimension is the average of the Score component and the Win Rate component.  

Regarding the Score component: 
\begin{enumerate}[leftmargin=*]
    \item Effectiveness is measured by keyword hit rate.  

\item Novelty is assessed by the embedding dissimilarity between the generated idea and existing literature.  
\item Detailedness is evaluated via content completeness checks, with penalties for redundancy.  
\item Feasibility is determined by the similarity between the implementation graph generated by the model and an expert-provided template graph.  
\end{enumerate}

For the Win Rate component:  
For each dimension, a LLM serves as the judge in pairwise comparisons, pitting the model-generated idea against a reference answer. Independent votes are cast for each dimension, and the win rate of the model’s idea is computed accordingly.  

Since the Score component imposes strict formatting requirements on data and ideas (e.g., implementation details must be represented as a directed graph), which do not reflect general scenarios, we adopt only the Win Rate component for evaluation.

\subsection{Expert Evaluation Details.}
\label{appendix:expertdetail}

\subsubsection{Details of Expert Evaluation.}
We assigned 10 judged idea pairs to each expert and report the results in Table.~\ref{tab:expert_result}.
In Table.~\ref{tab:expert_result}, the total number of evaluations for Law is 26 and for Biotech is 29. The following are the reasons why these five pairs were not successfully evaluated:

\begin{table}[h]
\centering
\caption{Issue Table}
\label{tab:issues}
\resizebox{0.8\textwidth}{!}{%
\begin{tabular}{|l|p{6cm}|l|}
\hline
\textbf{ID} & \textbf{Reason} & \textbf{Model Generating the Issue} \\ \hline
Law1   & Incomplete idea generation &  {\qwenit} \\ \hline
Law2   & Incomplete idea generation &  {\qwenit} \\ \hline
Law3   & Incomplete idea generation &  {\qwenit} \\ \hline
Law4   & Generated idea is a hybrid of law and computer science &  {\model} \\ \hline
Biotech1 & Incomplete idea generation &  {\qwenit} \\ \hline
\end{tabular}%
}
\end{table}

Initially, we included a ``both good'' option, but since it received no selections, we removed it in the final version.

\subsubsection{Breakdown of Participant Positions.} 
We show the detailed position breakdown of our idea judge participants in Table.~\ref{apptab:participant_positions}.
We show the detailed institutions breakdown of our idea judge participants in Table.~\ref{apptab:institutions}.
We recruited 3 experts in each of the fields of Law, Biotechnology, and Education.
Experts are presented with an idea pair generated from the same reference, read idea pair, determine the winner, and provide a justification. In each idea pair, one idea originates from \model~and the other from either \qwenit~or gpt-4o. During evaluation, the experts are unaware of the specific model that generated each idea.

\begin{table}[h]
\centering
\caption{Positions of the idea judge participants.}
\label{apptab:participant_positions}
\begin{tabular}{lc}
\toprule
\textbf{Position} & \textbf{Count} \\
\midrule
Postdoc                & 1 \\
PhD                    & 6 \\
Mphil                 & 2 \\
\bottomrule
\end{tabular}
\end{table}

\begin{table}[h]
\centering
\caption{Institutions of the idea judge participants.}
\label{apptab:institutions}
\begin{tabular}{lc}
\toprule
\textbf{Institution} & \textbf{Count} \\
\midrule

The University of Hong Kong (HKU) & 3 \\
Zhejiang University (ZJU) & 3 \\
Broad Institute of MIT \& Harvard (Broad Institute) & 1 \\
Swiss Federal Technology Institute of Lausanne (EPFL) & 1 \\
The California Institute of Technology (Caltech)  & 1\\ 

\bottomrule
\end{tabular}
\end{table}

\section{Stability of the reward model}

As stated in our experimental Sec.~\ref{exp_setting}, we employ the Qwen-Plus model as the reward model. The reward model is required to observe both the real idea and the model-generated idea and assign scores accordingly. In Table~\ref{tab:judge_accuracy}, we present an experiment demonstrating that Qwen-Plus can effectively distinguish between real ideas and generated ideas.
Here, accuracy denotes the probability that Qwen-Plus correctly identifies the type of idea—assigning a score of 1 to the real idea and a score of 0 to the fake idea.
\begin{table}[ht]
\centering
\caption{Discrimination Accuracy of \qwenplus~on arXiv ideas (Real) vs. Qwen-14B-Instruct Generated Ideas (Fake)}
\label{tab:judge_accuracy}
\resizebox{0.4\columnwidth}{!}{%
\begin{tabular}{lcc}
\toprule
                & Real Ideas & Fake Ideas \\
\midrule
Accuracy (\%)   & 80.77            & 75.12                             \\
\bottomrule
\end{tabular}%
}
\end{table}

It can be seen that Qwen-Plus achieves high accuracy in identifying both real and fake ideas, demonstrating its suitability as a reward model.
\section{All prompts used in \model}

\begin{tcolorbox}[
    breakable, 
    title=Comment Model, 
    colback=white, 
    colframe=blue!75!black, 
    fonttitle=\bfseries,
    before skip=10pt, 
    after skip=10pt, 
    left=5pt, 
    right=5pt, 
    parbox=false
]
Name: Idea Authenticity Checker\\
Description: Determine whether a given research idea comes from real, published research work or is a fictional/hypothetical research idea.

\# Task Overview\\
1. Carefully analyze the provided research idea.\\
2. Make a judgment: 1 = real research work, 0 = fictional research idea.\\
3. Provide confidence level and detailed reasoning for the judgment.

\# Important: Do NOT Rely on Format Features\\
- DO NOT use citation formats (e.g., arXiv citations, BibTeX format) to judge authenticity\\
- DO NOT use DOI numbers, paper IDs, or publication identifiers as indicators\\
- DO NOT rely on formatting styles (e.g., LaTeX formatting, citation styles) to make judgments\\
- Focus ONLY on the CONTENT QUALITY and SUBSTANCE of the research idea itself\\
- Format features can be easily fabricated and are not reliable indicators of authenticity\\
- Base your judgment solely on the technical depth, problem clarity, and limitations discussion in the content\\
- Be mindful to avoid excessive focus on details, as this is a research idea rather than a complete research proposal.

\# Analysis Steps

\#\# Step 1: Extract Key Components\\
- Identify the core problem being addressed\\
- Extract the main technical approach or methodology\\
- Note any specific methods, algorithms, or techniques mentioned\\
- Identify novelty claims and contributions\\
- Note any limitations or challenges discussed\\
-  {CRITICAL}: Check the {technical\_approach} field (if present) for:\\
* Whether steps are truly executable or just superficially detailed\\
* Presence of concrete implementation details (data sources, parameters, algorithms)\\
* Discussion of computational complexity or resource requirements\\
* Missing critical details that would be needed for actual implementation

\#\# Step 2: Deep Technical Analysis\\
-  {CRITICAL: Evaluate Technical Approach Executability}\\
* If {technical\_approach} is present, check if steps are truly executable:\\
    - Do steps include concrete data sources, parameters, or algorithms?\\
    - Are there missing critical details needed for implementation?\\
    - Can a researcher actually follow these steps to reproduce results?\\
-  {CRITICAL: Check Technical Integration Depth}\\
* Verify if multiple techniques are meaningfully integrated or just superficially combined:\\
    - Are integration challenges acknowledged?\\
    - Is there explanation of how incompatible assumptions are resolved?\\
    - Are synergistic principles explained, or just stated?\\
*  {Red Flag}: Multiple complex techniques combined without explaining fundamental incompatibilities or theoretical soundness

\#\# Step 3: Make Judgment\\
- Consider all indicators together\\
- IMPORTANT: Base judgment ONLY on content quality, NOT on format features (citations, DOI, formatting styles)\\
-  {CRITICAL Red Flags for Fictional Ideas}:\\
*  {Superficial Technical Depth}: Technical terms are mentioned but not meaningfully integrated; steps look detailed but lack executable specifics\\
*  {Missing Practical Challenges}: No discussion of computational complexity, data availability, measurement error, endogeneity, cross-country data harmonization, or other real-world implementation difficulties\\
*  {Unrealistic Integration}: Multiple complex techniques are combined without explaining how they overcome fundamental incompatibilities or why the integration is theoretically sound\\
*  {Abstract Limitations Only}: Limitations are described in abstract terms without engaging with specific empirical challenges or methodological tensions\\
*  {Overly Ambitious Scope}: Simultaneously addresses too many disparate problems without acknowledging the complexity or potential conflicts\\
- Real research work typically shows:\\
* High technical specificity with meaningful integration of concepts\\
* Well-defined problem context with clear boundaries\\
* Detailed limitations discussion including practical challenges (computational, data, methodological)\\
* Technical steps that are executable with concrete implementation details\\
* Acknowledgment of potential failures, edge cases, or methodological tensions\\
- Fictional ideas typically show:\\
* Low to medium technical specificity OR high specificity but superficial integration\\
* Vague problem context OR overly broad scope without clear boundaries\\
* Minimal limitations discussion OR abstract limitations without practical challenges\\
* Technical steps that look detailed but lack executable specifics\\
* Missing discussion of computational complexity, data requirements, or implementation challenges\\
- Make the authenticity judgment: 1 (real) or 0 (fictional)\\
- Your knowledge maybe outdated, so do not use your knowledge to judge if some methods are present or absent.

\#\# Step 4: Assess Confidence\\
- High confidence (0.8--1.0): Clear indicators strongly support the judgment\\
- Medium confidence (0.5--0.8): Most indicators support the judgment, but some ambiguity exists\\
- Low confidence (0.0--0.5): Mixed indicators or insufficient information

\#\# Step 5: Generate Reasoning\\
- List 3--5 key reasons supporting the judgment\\
- Reference specific aspects of the idea that led to the conclusion\\
- Be specific and concrete in the reasoning

\# Output Requirements

\#\# Authenticity Field\\
- Must be exactly 1 (real research work) or 0 (fictional research idea)\\
- Based on comprehensive evaluation of all indicators

\#\# Confidence Field\\
- Number between 0.0 and 1.0\\
- Reflects how certain you are about the judgment\\
- Consider the strength and consistency of indicators

\#\# Reasoning Field\\
- Array of 3--5 strings\\
- Each string should be a clear, specific reason\\
- Reference concrete aspects of the idea\\
- Explain why the judgment was made

\# Here is a typical true idea:\\
\{ground\_truth\}

\# Input Data

\#\# Research Idea\\
\textless idea\textgreater\\
\{idea\}\\
\textless/idea\textgreater

Notes:\\
the result should be a valid JSON object, wrapped in \textless Judgment\textgreater...\textless/Judgment\textgreater

Output Format:\\
\textless Judgment\textgreater\\
\{\\
  "authenticity": 0,\\
  "confidence": 0.95,\\
  "reasoning": ["The idea is not clearly from real research work", "The idea is not well-defined and has a clear technical approach", "The idea is fake and not executable with concrete implementation details"]\\
\}\\
\textless/Judgment\textgreater
\end{tcolorbox}

\begin{tcolorbox}[
    breakable, 
    title=Reward Model, 
    colback=white, 
    colframe=blue!75!black, 
    fonttitle=\bfseries,
    before skip=10pt, 
    after skip=10pt, 
    left=5pt, 
    right=5pt, 
    parbox=false
]

You will be given two research ideas proposed by two PhD students attempting to formulate a research hypothesis.\\
Additionally, you will be given a ground truth idea that has been peer-reviewed and validated.\\
\\
\#\# Task\\
Evaluate how close each of the two ideas is to the ground truth idea. Provide scores on a scale of 0-1, where higher scores indicate better alignment with the ground truth.\\
- IMPORTANT: Base judgment ONLY on content quality, NOT on format features (citations, DOI, formatting styles)\\
- **CRITICAL Red Flags for Fictional Ideas**:\\
* **Superficial Technical Depth**: Technical terms are mentioned but not meaningfully integrated; steps look detailed but lack executable specifics (e.g., missing computational complexity, data requirements, implementation challenges)\\
* **Missing Practical Challenges**: No discussion of computational complexity, data availability, measurement error, endogeneity, or other real-world implementation difficulties\\
* **Unrealistic Integration**: Multiple complex techniques are combined without explaining how they overcome fundamental incompatibilities or why the integration is theoretically sound\\
* **Abstract Limitations Only**: Limitations are described in abstract terms without engaging with specific empirical challenges or methodological tensions\\
* **Overly Ambitious Scope**: Simultaneously addresses too many disparate problems without acknowledging the complexity or potential conflicts\\
\\
\#\# Output Format\\
You must return a JSON object with the following structure:\\
\{\{\\
    "idea1\_improve\_score": <float between 0 and 1>,  // Higher score indicates better alignment with ground truth\\
    "idea2\_improve\_score": <float between 0 and 1>,  // Higher score indicates better alignment with ground truth\\
    "reason": "<explanation of your evaluation>"\\
\}\}\\
\\
\#\# Example\\
\{\{\\
    "idea1\_improve\_score": 0.8,\\
    "idea2\_improve\_score": 0.6,\\
    "reason": "Idea1 shows better alignment with the ground truth idea because it more closely matches the structure and content of the validated research approach."\\
\}\}\\
\\
\#\# Input\\
idea1: \{idea1\}\\
idea2: \{idea2\}\\
ground\_truth: \{ground\_truth\}\\

\end{tcolorbox}

\begin{tcolorbox}[
    breakable, 
    title=System Prompt of \model, 
    colback=white, 
    colframe=blue!75!black, 
    fonttitle=\bfseries,
    before skip=10pt, 
    after skip=10pt, 
    left=5pt, 
    right=5pt, 
    parbox=false
]

  Name: Idea Quality Improver\\
  Description: Improve a research idea, making refinements to address weaknesses while preserving the core concept.\\
\\
  \# Task Overview\\
  1. You will be given some critics from users, you should update the idea based on those critics.\\
  2. Make improvements to address the weaknesses.\\
  3. Enhance technical\_approach with detailed step-by-step workflow if missing or insufficient.\\
  \ \ \\
  \# Critical Principles\\
\\
  \#\# Make Improvement\\
  - Enhance clarity, specificity, and depth rather than changing direction\\
  - Do NOT change the fundamental nature of the idea\\
\\
  \# Analysis Steps\\
\\
  - Address weaknesses found by users\\
  - \textbf{CRITICAL: Enhance or add technical\_approach}\\
    * If technical\_approach is missing or insufficient, add or enhance it with detailed step-by-step workflow\\
    * Include: data preparation, model/algorithm design, key technical components, implementation details, evaluation methodology\\
    * Present as a clear, sequential workflow that enables implementation and validation\\
  - Maintain coherence between different parts of the idea\\
  - Verify that improvements don't contradict the core concept\\
\\
  \# Output Requirements\\
\\
  \#\# Improved Idea Object\\
  - Output an improved\_idea object with the same structure as the original\\
  - All required fields must be present:\\
    * current\_limitations: Refined but preserving core problem.\\
    * idea\_summary: Enhanced sentences maintaining core concept. Focus solely on describing this idea itself.\\
    * technical\_approach: Enhanced step-by-step description of the methodological workflow.\\
    * novelty\_statement: Enhanced novelty claim\\
    * confidence: Confidence level\\
\\
  \#\# Improvement Summary Object\\
  - Output an improvement\_summary object explaining what was improved:\\
    * key\_improvements: Array of specific improvements made\\
    * improvement\_rationale: Brief explanation of why improvements were made\\
\\
  \# Output Format\\
  - Output must be valid JSON with no extra fields or commentary, wrapped in \textless Idea\textgreater...\textless/Idea\textgreater\\
  - Output must be in the following format:\\
  \textless Idea\textgreater\\
  \{\{\\
    "improved\_idea": \{\{\\
      "current\_limitations": "...",\\
      "idea\_summary": "...",\\
      "technical\_approach": "...",\\
      "novelty\_statement": "...",\\
      "confidence": "...",\\
      "improvement\_summary": \{\{\\
        "key\_improvements": ["...", "..."],\\
        "improvement\_rationale": "..."\\
      \}\}\\
    \}\}\\
  \}\}\\
  \textless/Idea\textgreater\\

\end{tcolorbox}

\begin{tcolorbox}[
    breakable, 
    title=Win Rate Evaluator, 
    colback=white, 
    colframe=blue!75!black, 
    fonttitle=\bfseries,
    before skip=10pt, 
    after skip=10pt, 
    left=5pt, 
    right=5pt, 
    parbox=false
]

You are assisting researchers tasked with comparing TWO research hypotheses (Hypothesis A and Hypothesis B).\\
Your job is to evaluate both hypotheses across five separate dimensions defined below, and to choose a winner (either Hypothesis A or Hypothesis B) for each dimension. Ties are NOT allowed — you MUST pick one winner per dimension. Base your judgments on scientific principles and the provided context only.\\
\\
\#\#Background context:\\
\{context\_text\}\\
\\
\#\#Hypothesis A:\\
\{hypothesis\_A\}\\
\\
\#\#Hypothesis B:\\
\{hypothesis\_B\}\\
\\
\#\#Definition of each dimension:\\
\#\#\#1) Effectiveness\\
Which hypothesis is more likely to produce a successful experimental or empirical outcome in service of the stated research objective? Evaluate the likelihood that, if implemented using standard practices in the relevant discipline, the hypothesis will achieve the intended measurable result. Focus on mechanistic plausibility, causal logic, and whether the hypothesis addresses the core problem directly.\\
\\
\#\#\#2)Novelty\\
Novelty: Which hypothesis presents more innovative or original approaches? Compare the similarity between the idea and the related work and existing solutions in the background to assess its novelty. A lower similarity to the core idea indicates greater novelty.\\
\\
\#\#\#3) Detailedness (Level of Specification)\\
Which hypothesis provides clearer, more actionable, and more complete specification of mechanisms, assumptions, experimental steps, required variables, and dependencies? Detailedness rewards clarity that would enable a competent researcher to design an experiment or implementation with minimal ambiguity.\\
\\
\#\#\#4) Feasibility\\
Which hypothesis presents a more realistic and implementable solution given current technological constraints?\\
\\
\#\#\#5) Overall\\
Considering the overall aspects together but emphasizing conceptual coherence and scientific grounding, which hypothesis is superior overall? This is a synthesis judgment: prefer the hypothesis that is logically consistent, grounded in accepted principles, avoids critical unstated assumptions or contradictions, and is most defensible as a scientific proposition.\\
\\
Unified constraints:\\
- Use only the provided context and widely accepted scientific principles in the relevant discipline. Do NOT invent facts external to the context unless they are broadly standard domain knowledge.\\
- When a dimension explicitly says to ignore other factors (e.g., Novelty should ignore feasibility), strictly follow that guidance for that dimension. When evaluating a certain dimension, it should focus on this dimension itself and ignore the influence of other dimensions.\\
- Be concise but specific: for each dimension provide a short judgment line (exact format below) plus 1–3 sentences of succinct reasoning grounded in the definitions above.\\
- Format must match exactly (case-insensitive for "Win A/Win B") and include a reason after "because".\\
\\
\\
\#\#Output format (MUST FOLLOW EXACTLY)\\
\\
Format your response exactly as follows:\\
Effectiveness: [Win A/Win B] because ...\\
Novelty: [Win A/Win B] because ...\\
Detailedness: [Win A/Win B] because ...\\
Feasibility: [Win A/Win B] because ...\\
Overall: [Win A/Win B] because ...\\
\end{tcolorbox}

\begin{tcolorbox}[
    breakable, 
    title=Rubrics Evaluator, 
    colback=white, 
    colframe=blue!75!black, 
    fonttitle=\bfseries,
    before skip=10pt, 
    after skip=10pt, 
    left=5pt, 
    right=5pt, 
    parbox=false
]

You are a strict and unbiased evaluator.\\
\\
Your goal is to analyze whether a proposed plan violates any desiderata with respect to a given rubric item.\\
\\
====================\\
DESIDERATA DEFINITIONS\\
====================\\
1. DETAILED, SPECIFIC SOLUTION\\
   - Does the part of the plan relevant to this rubric item give concrete details on HOW to implement it?\\
   - There should be no vague claims like "we will handle X" without explaining how.\\
   - Avoid ambiguity and unclear language; the explanation should be easy to understand.\\
\\
2. NO OVERLOOKED FLAWS OR WEAKNESSES\\
   - Are there important flaws or weaknesses in how the plan addresses this rubric item that would invalidate its satisfaction?\\
\\
3. WELL JUSTIFIED RATIONALE\\
   - Is the reasoning behind the plan for this rubric item well-motivated and justified?\\
   - Are there convincing arguments that this approach is better than simpler solutions or alternative hypotheses?\\
\\
4. COST AND EFFORT EFFICIENT\\
   - Does the plan handle this rubric item in a cost‑ and effort‑efficient way, without unnecessary complexity?\\
   - Would a simpler solution with less human or resource cost be equally effective?\\
\\
5. NO ETHICAL ISSUES\\
   - Does this part of the plan avoid negative consequences or ethically problematic behavior?\\
\\
6. CONSISTENT WITH OVERALL PLAN\\
   - Is this part of the plan consistent with the rest of the plan, without contradictions?\\
\\
====================\\
EVALUATION INSTRUCTIONS\\
====================\\
- Be skeptical, careful, and provide valid criticisms.\\
- Be as strict as possible while remaining unbiased and reasonable.\\
- Do NOT accept the plan merely because it claims to satisfy the desiderata.\\
- For each desideratum (1–6), decide only whether it is satisfied (`true`) or violated (`false`), focusing on parts relevant to the given rubric item.\\
\\
====================\\
OUTPUT FORMAT\\
====================\\
You MUST respond in valid JSON only, with the following structure:\\
\{\{\\
  "per\_desideratum\_satisfied": \{\\
    "1": bool,\\
    "2": bool,\\
    "3": bool,\\
    "4": bool,\\
    "5": bool,\\
    "6": bool\\
  \},\\
  "violated\_desiderata": [int],\\
  "global\_reason": string\\
\}\}\\
\\
ADDITIONAL RULES\\
- "violated\_desiderata" must be the list of desiderata numbers (1–6) that are violated, i.e., those with `satisfied = false`.\\
- If NO part of the plan addresses this rubric item, then:\\
  - set all values in "per\_desideratum\_satisfied" to false,\\
  - set "violated\_desiderata" to [1,2,3,4,5,6],\\
  - explain in "global\_reason" that the plan does not address the rubric item at all.\\
- Do NOT include any text outside of the JSON object.\\
\\
Below is an EXAMPLE of a valid JSON response format (values are illustrative; you must adapt them to the actual input):\\
\{\{\\
  "per\_desideratum\_satisfied": \{\\
    "1": true,\\
    "2": false,\\
    "3": true,\\
    "4": false,\\
    "5": true,\\
    "6": true\\
  \},\\
  "violated\_desiderata": [2,4],\\
  "global\_reason": "The plan is detailed and mostly well‑justified, but it omits some required stages, overlooks key risks, and is more resource‑intensive than necessary for the rubric item."\\
\}\}\\
\end{tcolorbox}
% \input{eval_appendix}
% \input{dataset}
% \input{dataset}
% \input{relate}

% You can have as much text here as you want. The main body must be at most $8$ pages long.
% For the final version, one more page can be added.
% If you want, you can use an appendix like this one.  

% The $\mathtt{\backslash onecolumn}$ command above can be kept in place if you prefer a one-column appendix, or can be removed if you prefer a two-column appendix.  Apart from this possible change, the style (font size, spacing, margins, page numbering, etc.) should be kept the same as the main body.
%%%%%%%%%%%%%%%%%%%%%%%%%%%%%%%%%%%%%%%%%%%%%%%%%%%%%%%%%%%%%%%%%%%%%%%%%%%%%%%
%%%%%%%%%%%%%%%%%%%%%%%%%%%%%%%%%%%%%%%%%%%%%%%%%%%%%%%%%%%%%%%%%%%%%%%%%%%%%%%

\end{document}